\documentclass[twocolumn,a4paper]{article}
\usepackage[margin=0.75in]{geometry}

\def\volumeyear{2019}

\usepackage{graphicx}        
\usepackage[lined,ruled]{algorithm2e}
\usepackage{tabularx}
\usepackage{amsmath}
\usepackage{txfonts}
\usepackage{subfig}
\usepackage[export]{adjustbox}
\usepackage{xcolor}
\usepackage{breqn}

\newcommand{\sjgComment}[1]{\textcolor{orange}{#1}}
\newcommand{\bcomment}[1]{\textcolor{blue}{#1}}
\newcommand{\icomment}[1]{\textcolor{red}{#1}}
\newcommand{\iicomment}[1]{\textcolor{magenta}{#1}}
\newcommand{\IGNORE}[1]{}
\usepackage{enumitem}

\usepackage{natbib}
\usepackage{hyperref}

\setlist{nolistsep}

\begin{document}

\title{NH-TTC: A gradient-based framework for generalized anticipatory collision avoidance}

\author{Bobby Davis$^1$, Ioannis Karamouzas$^2$, and Stephen J. Guy$^1$}
\date{\small
	${}^1$Department of Computer Science and Engineering, University of Minnesota \\
	${}^2$School of Computing, Clemson University
}



\maketitle

\begin{abstract}
We propose NH-TTC, a general
method for fast, anticipatory collision avoidance 
for autonomous robots having arbitrary equations of motions.
Our proposed approach exploits implicit differentiation and subgradient descent to locally  optimize  the  non-convex and non-smooth cost  functions  that  arise  from  planning  over  the anticipated  future positions of nearby obstacles. The result is a flexible framework capable of supporting high-quality, collision-free navigation with a wide variety of robot motion models in 
various challenging scenarios. We show results for different navigating tasks, with our method controlling various numbers of agents (with and without reciprocity), on both physical differential drive robots, and simulated robots with different motion models and kinematic and dynamic constraints, including acceleration-controlled agents, differential-drive agents, and smooth car-like agents.
The resulting paths are high quality and collision-free, while needing only a few milliseconds of computation as part of an integrated sense-plan-act navigation loop.  The associated video is available at \url{http://motion.cs.umn.edu/r/NH-TTC}{}.
\end{abstract}


\section{Introduction}

Recent trends in robotics, machine learning, and computer graphics have significantly advanced the state-of-the-art in autonomous navigation of mobile robots and intelligent agents. Over the past two decades, numerous \emph{global} approaches have been proposed including  incremental sampling-based planners~\citep{rrt*,lavalle2001randomized} and 
receding horizon
formulations~\citep{falcone2007predictive,information} which compute a sequence of controls for the robot by optimizing a cost function that accounts for goal-oriented safe navigation. 
In many situations, though, an important task for the robot is to \emph{immediately react} to its local surroundings while still making its best effort to follow its global plan. Whether it is an autonomous car driving on a highway, or a semi-autonomous smart-shelf navigating in an automated warehouse, a robot should be able to observe its surroundings, \emph{anticipate} the expected behavior of nearby obstacles, and react accordingly, all within a tight sense-plan-act loop. 

Many successful recent techniques for anticipatory local navigation have been based on the concepts of velocity obstacles (VO)~\citep{FS98,rvo}, that is, the set of all possible relative velocities that will lead to a collision between two robots. VO-based formulations of collision avoidance are naturally anticipatory in that they penalize upcoming, future collisions. Recent work has extended the basic idea of VOs to allow efficient real-time solutions for complex scenarios  ~\citep{clearPath},  provide formal guarantees of collision avoidance with multiple interacting agents~\citep{orca}, and shown how to support a variety of non-holonomic motion models~\citep{gvo,gorca}. Broadly speaking, these methods work by computing 
(conservative) linear approximations of the complex action spaces that are known to be collision free with respect to a given dynamic obstacle. Optimal controls can then be found using geometric optimization techniques, leading to an overall approach that is computationally efficient, but can be overly conservative, especially in complex scenarios.

Inspired by the success of these geometric optimization techniques, we seek to propose a more generalized framework for representing, specifying, and accounting for the anticipatory collision avoidance needs of mobile robots. 
While previous local methods typically plan for robot actions in some intermediate space (e.g., a velocity space), we propose a method that allows robots to quickly plan directly in the space of their controls. This allows us to naturally support planning over a wide variety of robot types with various kinematic or dynamic constraints. 

Our NH-TTC approach, short for non-holonomic time-to-collision,
leverages gradient-based optimization techniques to develop an ``anytime" framework for iteratively refining a robot's control to minimize the expected costs of the resulting trajectories. We achieve this by
borrowing techniques from trajectory optimization. Specifically, we compose a cost function consisting of two terms: a time-to-collision based trajectory cost and a cost-to-goal term. In addition, to enable quick optimizations, we consider only trajectories created by a single, constant control.
The result is a high quality estimate of the true cost of a control that can still be quickly optimized even for complex, non-holonomic dynamics models with kinematic constraints.


Two key technical challenges must be addressed to enable our approach:
\begin{enumerate}
    \item 
    Because the new cost term we introduce is a function of the expected future states of the robot, it will have  discontinuities in its values (e.g., an arbitrarily small change in control can move a robot discontinuously into a collision course). 
    \item There is typically no closed-form, analytical solution that can represent the dynamics of future collision that will develop over time for arbitrary robot motion models.
\end{enumerate}
Together, these two issues can prevent the use of many traditional gradient-based techniques for finding optimal controls. The lack of a closed-form solution means we can not easily compute the true gradient of the cost function with respect to the robot controls. In addition, the discontinuity in the cost means that even if we can estimate this gradient well for parts of the solution space, it will be undefined at the discontinuities.

In this paper, we propose NH-TTC, a generalized framework for fast, anticipatory, optimization-based steering that addresses the above challenges by exploiting \emph{implicit differentiation} and \emph{subgradient descent} to locally optimize the nonconvex and nondifferentiable cost functions that arise from planning over expected future positions of neighboring obstacles. We show the applicability of our solution both on physical robots and simulated agents including smooth differential drives, double integrators, and car-like agents. Moreover, the method can be easily extended to more complicated navigation tasks such as chasing dynamic goals and multi-robot navigation in shared environments.

Our proposed framework occupies an important middle ground between reactive methods that only account for local state and planning-based approaches which offer full trajectory optimization. We can provide fast computation appropriate for use in a closed-loop, reactive control scheme, while still providing high-quality paths that avoid the long-term inefficiencies of purely local optimization.

\textbf{Organization.} The remainder of the paper is organized as follows. Section~\ref{sec:rel_work} reviews relevant work on decentralized planning and local collision avoidance. 
Section~\ref{sec:overview} casts local navigation as a control optimization problem, and Section~\ref{sec:control_opt} details our generalized optimization framework that relies on subgradient descent and implicit differentiation. 
Section~\ref{sec:dynamics_models} describes how our method can be applied to different dynamics models, including a smooth car, and Section~\ref{sec:implementation} provides details regarding our experimental setup and the implementation of NH-TTC.
In Section~\ref{sec:collision_avoidance}, we show the applicability of NH-TTC to simulated agents 
navigating amidst dynamic obstacles. 
Section~\ref{sec:extensions} shows how our approach can be extended to support dynamic goals, and to enable collision avoidance in a multi-robot setting.
Section~\ref{sec:analysis_experiments} presents results on physical robots and analyzes the performance of NH-TTC. 
Finally, in Section~\ref{sec:conclusion}, we conclude, and discuss limitations along with future research directions.

\section{Related Work} \label{sec:rel_work}
There is an extensive amount of literature on planning global trajectories among static and moving obstacles, including sampling-based approaches~\citep{rrt*,jaillet2004prm}, uncertainty-aware planners~\citep{belieftrees,lqgmp}, anytime and receding horizon approaches~\citep{falcone2007predictive,mattingley2011receding} and more recently end-to-end deep learning techniques~\citep{agile}, to name just a few. 
Given the current focus of our work on realtime mobile robot navigation, in the rest of this section, we restrict ourselves to some highly relevant work on decentralized planning and local collision avoidance. 

Our method falls under the class of \emph{anticipatory collision avoidance} methods in which the robot is able to predict how its neighborhood evolves over time and react accordingly~\citep{dynamic}. 
Most of such anticipatory methods rely on the concept of Velocity Obstacles (VO) introduced by \citet{FS98} that defines the set of all relative velocities that will lead to a collision between a holonomic robot and moving obstacles at some moment in time. VOs provide a tractable alternative to inevitable collision states~\citep{ics,ics2}, and have been successfully extended to account for reciprocity between agents allowing their application to decentralized multiagent navigation problems~\citep{rvo,clearPath}. Further extensions have been proposed including VO definitions for reciprocal collision avoidance between robots having more complicated dynamics~\citep{avo,cco,lqr}, multi-robot teams walking in formation~\citep{coherence,fvo}, and formulations that account for uncertainty in the future trajectory of obstacles~\citep{pvo,hrvo}.

The Generalized Velocity Obstacles (GVO) approach was introduced by~\citet{gvo} that defines collisions with obstacles in the control space and enables safe navigation of kinematically constrained robots by selecting a control outside the ``control obstacles" space. Our approach is closely related to GVOs, as we also plan directly in the control space of the robots. However, GVO relies on the discretization of the control space, which can become prohibitively expensive for real-time implementation on robots with complex dynamics, and also assumes linear motion for the obstacles. In contrast, we use numerical optimization methods to search for a local optimum control in an anytime fashion and can support different motion models for the obstacles. 

To address the challenges that sampling-based VOs pose for real-time multi-robot navigation, the popular ORCA framework was proposed by~\citet{orca}. ORCA conservatively approximates VOs as half-planes, allowing robots to quickly find collision-free velocities outside of the union of all VOs by solving a convex optimization problem through linear programming. Since the original work of van den Berg et al. on holonomic robots, many ORCA-based approaches have been proposed that linearize the VOs or learn such constraints,  including approaches for steering differential-drives, car-like robots, and other non-holonomic agents~\citep{rrvo,orcadd,mora1,mora2,jia1}. 

More closely related to our work is the Generalized Reciprocal Velocity Obstacles (GRVO) approach of~\cite{gorca} that provides a generalized framework for local navigation supporting robots with both linear and nonlinear equations of motions.  Similar to GVO, GRVO plans over potential robot controls, but does so through an indirect manner by representing a set of controls over time as a single high-level target velocity. GRVO then ``ORCAfies"  this space of target velocity using linear approximations to represent which target velocities may lead to collisions.  
However, GRVO, and other ORCA-like approaches, can be overly conservative in their approximations, forbidding large amounts of admissible controls due to the linearization of control constraints. 
Furthermore, these approaches all typically use a binary indicator cost function to assess the quality of a given control input; a control is either strictly forbidden or fully allowed. Together, these factors can lead to inefficient robot behavior, especially in highly constrained scenarios.

To address these issues, we propose an alternative, gradient-based, framework, where the robot selects a new control by directly optimizing an anticipatory cost function inspired by recent work in understanding collision avoidance between pedestrians~\citep{prl,mpd}. We note that while gradient-based steering has been explored before for local navigation~\citep{uttc,cgf17}, such approaches focus only on holonomic agents. Importantly, most of the existing work in anticipatory local navigation, including recent socially-inspired approaches and probabilistic-based frameworks~\citep{braids,warprss}, plans in an intermediate, higher-level space (such as velocities).
In contrast, we focus on reactive local navigation directly in the control space.
Here we assume that a robot can interact with other agents that \emph{may or may not react} to it, and propose a generalized local steering approach that can be applied to a variety of robot motion models and navigation settings.

\section{NH-TTC Problem Formulation} \label{sec:overview}
Our work considers the problem of a robot that must traverse among moving obstacles while navigating to a goal position. Here, we formulate the problem as one where the robot is following a tight sense-plan-act loop many times a second. As such, our approach is similar to classic  ``reactive" planning approaches as the robot is given only a few milliseconds to compute new controls each time step in response to its immediate sensor input. Existing reactive approaches typically have one of two key limitations: either they need to use simple dynamics models for the robots (e.g., ORCA assumes direct velocity control), or they work in some intermediate representation (e.g., positional fields or velocity-space) that must be translated somehow to robot controls. 
In contrast, we directly compute an exact control that is (locally) optimal with respect to some cost function, which allows both the robot and the obstacles in its environment to have (different) arbitrary dynamics functions. 


Our approach has two key features that enable strong practical performance. First, the approach is anytime, this means that it can quickly find an acceptable solution (typically, well under a millisecond)  
and iteratively refines it as time is available. Secondly, as in full trajectory optimization approaches, our approach is anticipatory, penalizing controls based on their expected future impact.  Unlike in full trajectory optimization approaches, we focus on trajectories generated by a single control, allowing NH-TTC to quickly find good paths.

\subsection{Notation and Preliminaries}
We assume our environment contains a single robot navigating to a goal position $\mathbf{p}_g$  while avoiding a set of obstacles 
(see Section~\ref{sec:reciprocity} for the application of our approach to multi-agent settings with several robots navigating simultaneously in a shared environment).
We assume both the robot itself and the various obstacles in its environment follow some known continuous time dynamics functions that defines the future state of the robot $\mathbf{x}(t,\mathbf{u})$ and obstacles states $\mathcal{O}(t) = \{ \mathbf{o}_i(t) ,\ \forall i\}$ 
as follows:
\begin{equation}
\begin{split}
\dot{\mathbf{x}}(t,\mathbf{u})&=f(\mathbf{x}(t,\mathbf{u}),\mathbf{u}) \\
\dot{\mathbf{o}}_i(t)&=g_i(\mathbf{o}_i(t)),
\end{split}
\end{equation}
where $\mathbf{u}\in\mathcal{U}$ is a valid control input, and $f$, $g$ are (possibly non-linear) continuous-time equations of motion. The set $\mathcal{U}$ is used to encode constraints on the robots dynamics, such as maximum control limits. Likewise, we can define a collection of valid states $\mathcal{X}$ that can be used to constrain any aspect of the robot's state such that $\mathbf{x}(\mathbf{u},t)\in\mathcal{X}~,\forall t$. This is needed to specify state constraints that are not directly part of a robot's control. For example, an acceleration-controlled robot may have a maximum velocity.

To determine collisions between the robot and the obstacles, we model them both as disks.  These disks are defined by projecting both the robot and obstacle states into a common Euclidean workspace, typically 2d or 3d, and then finding the minimal covering disk.  As such, we 
define the robot's position as 
$\bar{\mathbf{x}}(t,\mathbf{u}) = p(\mathbf{x}(t,\mathbf{u}))$, where $p$ maps from state space to the Euclidean workspace. 
The function $p$ is chosen to  place the center of the collision disk with a radius $r_x$ so as to wrap the true shape of the robot as closely as possible.
%
Similarly, let $\bar{\mathbf{o}}_i(t)=q_i(\mathbf{o}_i(t))$ and $r_{o_i}$ be the center of the collision disk and its radius for obstacle $i$ at time $t$, respectively, where the function $q_i$ maps the obstacle's state space to the workspace.

\subsection{Optimization-based Formulation}
Given the above notation, we can formally define the problem as follows.  We are given the robot's current state, $\mathbf{x}(0)$, a set of obstacle states over time, $\mathcal{O}(t)=\{\mathbf{o}_i(t),\ \forall i\}$, and the robot's goal position, $\mathbf{p}_{goal}$, which we assume is been computed by a high level planning approach.  
The task for the robot is find a collision-free trajectory, $\mathbf{x}(t)$, that approaches the goal as fast as possible while obeying the constraints of the robot dynamics, $\dot{\mathbf{x}}(t,\mathbf{u})=f(\mathbf{x}(t,\mathbf{u}),\mathbf{u})$, the control constraint set, $\mathcal{U}$, and accounting for the state constraint set $\mathcal{X}$.

Given an arbitrary trajectory $\mathcal{T} = \{\mathbf{x}(t),\ \forall t\geq0\}$ we seek to construct a cost function $C(\mathcal{T})$ which evaluates how well the trajectory does at providing efficient, collision-free motion towards the robot's current goal. Similar to other trajectory optimization approaches we break this cost into two terms:
\begin{equation}
    C(\mathcal{T})=C_{goal}(\mathcal{T}) + C_{col}(\mathcal{T})
    \label{eq:cost_func_traj}
\end{equation}
where $C_{goal}(\mathcal{T})$ evaluates how closely  the trajectory comes to meeting its goal state (or goal position), and $C_{col}(\mathcal{T})$ assigns a penalty to trajectories which have a high risk of collision.  Given sufficient computation time, our goal would be to find the complete trajectory which minimizes this cost function, e.g., via sampling as in \cite{bekris2007greedy}, \cite{rrt*}, and \cite{denny2013rrt}, or using a POMDP-like formulation as in \cite{Platt2010BeliefSP} and \cite{MacDecPOMDP}. However, these methods typically take several seconds or longer to converge so are inappropriate for the real-time setting considered here.

In order to allow the fast computation needed for use in a tight reactive planning loop, we consider only trajectories that are represented as a single, consistent, control $\mathbf{u}$ that is executed indefinitely. As a result, we can reparameterize our cost function in terms of a single control $\mathbf{u}$:
\begin{equation}
    C(\mathbf{u})=C_{goal}(\mathbf{u}) + C_{col}(\mathbf{u})
    \label{eq:cost_func}
\end{equation}
and reformulate the task to one of finding the optimal control. That is, finding the single control that leads to the optimal fixed-control trajectory as defined by Equation~\ref{eq:cost_func}.

In practice, a robot would not take this fixed-control trajectory indefinitely. Rather, it will re-run this optimization many times a second updating its planned trajectory as it approaches the goal and as local conditions change (similar in spirit to a receding horizon planner). Additionally, robots typically have some limits on what controls they make take at a given instant, such as maximum accelerations or steering limits, which leads to a set of admissible controls $\mathcal{U}$ at any time. The result is a constrained optimization problem
\begin{equation}
\begin{aligned}
    &\min_{\mathbf{u}} && C(\mathbf{u})\\
    &\mathrm{Such\ that:} && \mathbf{u}\in\mathcal{U}\\
    &&&\mathbf{x}(t,\mathbf{u})\in\mathcal{X} \quad \forall t\geq 0
    \label{eq_optimize}
\end{aligned}
\end{equation}
that the robot must solve at each step of its sense-plan-act loop. To ensure state constraints are satisfied, we must also check if the resulting trajectory respects the state constraints $\mathcal{X}$, and, if not, project $\mathbf{u}$ to the nearest control that respects these constraints within a small time horizon.

\subsection{NH-TTC Cost Function}

\begin{figure}
    \centering
    \captionsetup[subfigure]{justification=centering}
    \subfloat[Scenario]{
    \fbox{\includegraphics[width=0.9\columnwidth]{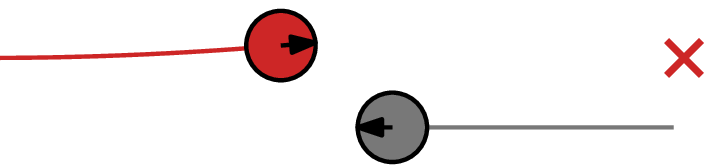}}}
    \\
    \subfloat[Cost Field]{\includegraphics[width=.9\columnwidth]{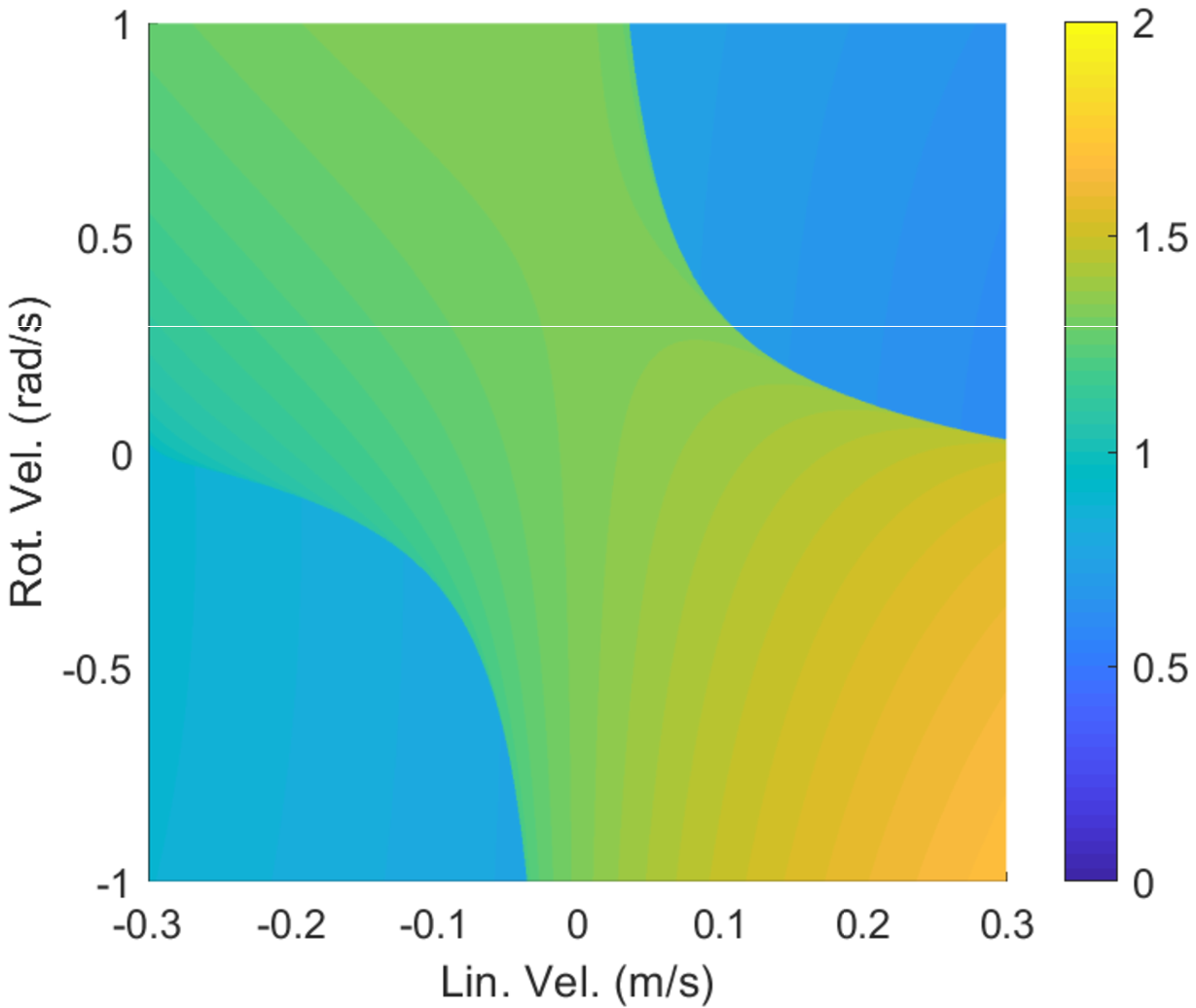}}
    \caption{
    \textbf{Gradient-Based Optimization in Control Space}\\ (a) A differential-drive robot, shown in red, has to reach the `x' mark while avoiding
    a non-reactive obstacle shown in gray. 
(b) The corresponding cost field of the robot is visualized here by taking samples from the robot's feasible controls. To better show the gradient, values are plotted in log scale with the color corresponding to $\log(C(\mathbf{u}) + 1)$.
}
\label{fig:1v1_energy}
\end{figure}

In general, many different cost fields can be considered in Equation~\ref{eq:cost_func}.
Inspired by trajectory optimization techniques, we find that the sum of two simple cost functions works well in a wide variety of scenarios: a cost-to-goal and a time-to-collision based trajectory cost.

\paragraph{Goal Cost ($C_{goal}$):}
Because a fixed-control trajectory has no end point, we evaluate the position of the robot at some time $t_{goal}$ into the future. The goal cost is then defined by 
how close the robot will be to its goal position at that time:  
\begin{equation}
    C_{goal}(\mathbf{u}) = \kappa_{goal}\ \left\Vert
    \bar{\mathbf{x}}(t_{goal},\mathbf{u})-\mathbf{p}_g \right\Vert,
\label{eq:goal_cost}
\end{equation}
where $\kappa_{goal}$ is a scaling constant. 
Section~\ref{sec:dyn_goals} discusses an extension to this cost function to better support cases where a robot is chasing after a moving target.

\paragraph{Collision Cost ($C_{col}$).}
Again, as our fixed-control trajectory has no endpoint, we evaluate collisions only for the next $t_{horiz}$ seconds. Inspired by recent findings that suggest the urgency humans place on a collision 
follows an inverse power-law relationship with how imminent such a collision is
\citep{prl}, we penalize trajectories with a term that is inversely proportional to the time until the nearest collision:
\begin{equation}
    C_{col}(\mathbf{u}) = \max_{\mathbf{o}_i\in \mathcal{O}}\frac{\kappa_{col}}{\tau(\mathbf{u}, \mathbf{o}_i)}
    \label{eq:col_cost}
\end{equation}
Here $\kappa_{col}$ is a scaling constant and $\tau(\mathbf{u}, \mathbf{o}_i)$ computes the minimum time to collision between the robot's trajectory (as determined by the control, the current robot position, and the dynamics of the robot) and the expected future trajectory of obstacle $\mathbf{o}_i$.  
The result of this human-inspired collision cost is a natural balance between strongly avoiding urgent collisions and (when necessary) taking controls that will lead to collision in the far enough in the future that the robot will have a chance to re-plan well before the collision happens. 
Note that, to improve efficiency, we only consider the first collision that happens within the next $t_{horiz}$ seconds.

An important property of our cost function is that it rises to infinity as the time until the nearest collision approaches zero. This means controls which lead to immediate collisions have effectively an infinite penalty. As a result, in the limit as planning frequency approaches infinity, our approach is guaranteed to be collision-free so long as the optimizer has sufficient time to converge to a finite cost (and assuming a collision-free control exists).

\subsection{Challenges}

Despite the straightforward nature of its components $C_{goal}$ and $C_{col}$, the resulting cost function $C$ is difficult to optimize as it is constrained, non-convex, and non-continuous.  Specifically, time to collision has a sharp discontinuity between a glancing collision and no collision, jumping from a finite value to infinity, respectively. As an example of these discontinuities, Figure~\ref{fig:1v1_energy} shows the cost field, $C$, for a robot avoiding a single (passive) oncoming obstacle. Here, the robot is assumed to be a differential-drive robot that can directly control its linear and angular velocity.

As can be seen in Figure~\ref{fig:1v1_energy}, there are sharp boundaries between the collision and collision-free controls. Controls that drive backwards (the left half of the figure) have higher costs due to the $C_{goal}$ term, controls that the lie along the top left or bottom right quadrants have addition penalties from $C_{col}$ as they will lead to imminent collisions. The optimal control (linear velocity = 0.3$\,$m/s, rotational velocity = 0.03$\,$rad/s) steers towards the goal, while gently avoiding any potential collisions.

Due to the discontinuities, classical gradient descent will be ineffective (as will other standard, higher order optimization techniques). Note that smoothing this cost field as in \cite{implicitTTC} is not possible with arbitrary dynamics, since we do not in general know a priori where the discontinuities lie.
Additionally, $C$ itself can be difficult to compute as we may have no closed form solution for the position of the robot or the obstacles at any given time depending on their dynamics model. The following section describes our approach to overcoming these difficulties.

\section{Control Optimization} \label{sec:control_opt}
Per our problem formulation, optimizing Equation~\ref{eq_optimize} will allow us to compute a per-time step control for the robot that will result in kinematically feasible, anticipatory, goal-oriented navigation in a constrained  settings. Our proposed NH-TTC approach 
relies on two techniques.
First, we will use subgradient descent-based optimization~\citep{subgradient,bertsekas1999nonlinear} which will allow us to find local minima even in the presence of a non-continuous, non-smooth cost function. Second, we will approximate the robot and obstacle dynamics via Runge-Kutta integration and compute the control gradients of the robot's position and time to collision through the use of \textit{implicit differentiation}.

\subsection{Subgradient Descent}
\label{subsec:control_opt:subgradient}
In order to optimize Equation~\ref{eq_optimize} we use the subgradient descent algorithm. As in standard gradient descent, the control $\mathbf{u}$ is iteratively updated based on the gradient of the cost with respect to $\mathbf{u}$:
\begin{equation}
\mathbf{u}_{k+1} = \mathbf{u}_{k} - \alpha \, \mathbf{s}_k
\end{equation}
where the search direction, $\mathbf{s}_k$, is based on the gradient $dC/d\mathbf{u}$.
However, at points where $C$ is not smooth, and therefore $dC/d\mathbf{u}$ is undefined,  we choose either the left or right gradient. This is known as the subgradient, which we will call $\mathbf{g}_k$.  While this gives us an optimization direction, it does not define the stepsize $\alpha$, i.e., how far to update our control in that direction. 

Numerous techniques have been proposed to choose an appropriate update size $\alpha$.  Experimentally, we found a Polyak update-based approach~\citep{polyak} to perform well in our domain. This method assumes the optimal possible control cost, $c^*$, is known in advance. Then, at each descent iteration $k$, we compute the difference between the current cost, $c_k$, and the optimal value of the function, $c^*$, and scale the result by the squared magnitude of the current subgradient search direction,~$\left\Vert\mathbf{s_k}\right\Vert^2$: 
\begin{equation}
\label{eq_poly}
\alpha = (c_k - c^*)/\left\Vert\mathbf{s}_k\right\Vert^2
\end{equation}
Given the complex and dynamic nature of our cost function, it is generally not possible to know the optimal value $c^*$ in advance. As such, we compute an approximate optimal cost $\hat{c}^*$ by first taking the best cost seen in any of the iterations so far, $c^+$, and then subtracting a small amount:
\begin{equation}
    \hat{c}^* = c^+ - \frac{10}{10 + k}
\end{equation}
As has been suggested in the literature~\citep{subgradient2,nedic2001incremental}, the amount we subtract from $c^+$ gradually decreases with increasing iteration count, $k$. The resulting update to $\alpha$ guarantees that the subgradient decent will converge on a true local minimum as $k$ approaches infinity.

While it is possible to directly use the subgradient as the search direction (i.e., $\mathbf{s}_k = \mathbf{g}_k$), we found convergence to be improved by adding ``momentum" to the search direction~\citep{nesterov2003introductory}. That is, our search direction at iteration $k$ is based in part on the current subgradient $\mathbf{g}_k$  and in part on the previous search direction, $\mathbf{s_{k-1}}$. 
Experimentally, we found the following update rule to work well:
\begin{equation}
    \mathbf{s}_k = \frac{1}{2}(\mathbf{s}_{k-1} + \mathbf{g}_k)
\end{equation}

Subgradient descent does not guarantee a cost decrease at each iteration, so the best control seen across the entire optimization is used as our final result.
Additionally, we project the control computed at each iteration onto the control constraints, $\mathcal{U}$, to ensure the controls remain feasible. The resulting 
projected subgradient descent algorithm is shown in full in Algorithm~\ref{alg:sgd}.  Here, we assume that the robot is given a time budget, $\mathrm{max\_time}$, to compute its new  control.

\begin{algorithm}[t]
\SetKwInOut{Input}{Input}
\SetKwInOut{Output}{Output}
\Input{$\mathbf{u}_0,\ \mathcal{U},\ \mathrm{max\_time}$}
\Output{$\mathbf{u}^*$}
$k=0$\\
$\mathbf{u}^* = \mathbf{u}_k = \mathbf{u}_0 $\\
$c^* = c_k = \mathcal{C}(\mathbf{u}_0) $\\
$\mathbf{s}_{k-1} = \mathbf{0}_{n\times 1}$\\

\While{$\mathrm{elapsed\ time} < \mathrm{max\_time}$}{
    $\mathbf{g}_k = dC / d\mathbf{u}(\mathbf{u}_k)$\\
    $\mathbf{s}_k = \frac{1}{2}(\mathbf{s}_{k-1} + \mathbf{g}_k)$\\
    $\hat{c}_k^* = c^* - 10 / (10 + k)$\\
    $\alpha = (c_k - \hat{c}_k^*)/\left\Vert \mathbf{s}_k \right\Vert^2$\\
    
    $\mathbf{u}_{k+1} = \mathbf{u}_k - \alpha\,\mathbf{s}_k$\\
    Project $\mathbf{u}_{k+1}$ into $\mathcal{U}$\\
    $c_{k+1} = \mathcal{C}(\mathbf{u}_{k+1}) $\\
    Update $c^*$ and $\mathbf{u}^*$\\
    $k = k+1$
    
}
\caption{Subgradient-Based Control Optimization}
\label{alg:sgd}
\end{algorithm}

\subsection{Subgradient Computation}
\label{sec:control_opt_computation}
To run the subgradient algorithm, we need to be able to compute the gradient, $\mathbf{g}_k = dC / d\mathbf{u}(\mathbf{u}_k)$.  This is non-trivial as there may not be a closed form solution for the robot position, the obstacle position, or the time to collision.
Below, we first discuss how to compute the cost function, and then focus on the computation of the gradient.


\subsubsection{Cost Computation}
To compute the goal cost, $C_{goal}$ (Equation~\ref{eq:goal_cost}), we need to compute the position at $t_{goal}$.  As we may not have a closed form solution for $\mathbf{x}(t_{goal},\mathbf{u})$, we approximate it using fourth order Runge-Kutta integration (RK4). To improve accuracy, we iteratively run multiple steps of RK4, such that each step is, at most, some small time horizon, $dt_{max}$.  This parameter allows for tuning the accuracy of our position estimation, at the expense of computation time.  Once $\mathbf{x}(t_{goal},\mathbf{u})$ has been computed, we pass it through the position mapping function $p$ to obtain the Euclidean position, $\bar{\mathbf{x}}(t_{goal},\mathbf{u})$, which, along with the goal position, fully defines the goal cost.

\begin{figure}
	\centering
	\null\hfill
	\includegraphics[width=0.7\columnwidth]{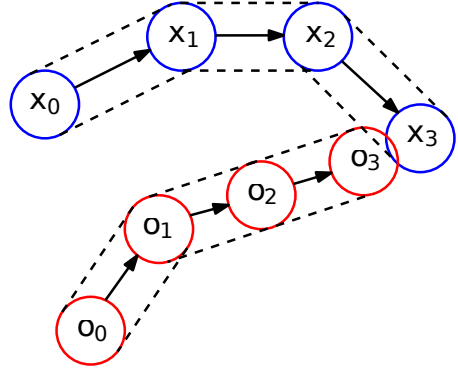}
	\hfill\null
	\caption{\textbf{Time to Collision Computation}: Robot and obstacle states are forward propagated using RK4 integration, and then linear continuous collision checks are done between each discrete state.}
	\label{fig:collision_check}
\end{figure}

To compute the collision cost, $C_{col}$ (Equation~\ref{eq:col_cost}), we need to compute the most imminent time to collision over all the obstacles.
Assuming both objects are approximately circular, the time to collision between an agent $\mathbf{x}$ and and object $\mathbf{o}$ can be defined as the time at which the two disks touch, i.e.:
\begin{equation}
\label{eq:ttc_char}
\left\Vert \bar{\mathbf{x}}(\tau(\mathbf{u},\mathbf{o}),\mathbf{u})) - \bar{\mathbf{o}}(\tau(\mathbf{u},\mathbf{o})) \right\Vert^2 - (r_x+r_{o})^2 = 0
\end{equation}
However, for many systems, solving this equation for $\tau$ is not feasible, as there may not be a closed form solution for $\bar{\mathbf{x}}$ and/or $\bar{\mathbf{o}}$, or the resulting equation is too complex.
Instead, we utilize a similar approach to that used to compute the goal cost.  
We forward propagate the state of the robot and the state of each obstacle using RK4, and perform linear continuous collision checks between the resulting states to estimate the first moment of collision that may have occurred during the integration steps (see Figure~\ref{fig:collision_check}).

Knowing how to compute the (propagated) cost $C$, we next show how to compute the gradient of the cost with respect to the controls $\mathbf{u}$.

\subsubsection{Goal Cost Gradient}
\label{sec:goal_deriv}
As Equation~\ref{eq:goal_cost} does not directly rely on $\mathbf{u}$, we must apply the chain rule to compute the gradient:
\begin{equation}
    \frac{dC_{goal}(\mathbf{u})}{d\mathbf{u}} =
    \frac{dC_{goal}(\mathbf{u})}{d\bar{\mathbf{x}}(t_{goal},\mathbf{u})}\frac{d\bar{\mathbf{x}}(t_{goal},\mathbf{u})}{d\mathbf{x}(t_{goal},\mathbf{u})}\frac{d\mathbf{x}(t_{goal},\mathbf{u})}{d\mathbf{u}}
\label{eq_dCdU}
\end{equation}
The first term can be directly computed as:
\begin{equation}
\frac{dC_{goal}(\mathbf{u})}{d\bar{\mathbf{x}}(t_{goal},\mathbf{u})} =
\frac{\kappa_{goal}}{2\left\Vert
    \bar{\mathbf{x}}(t_{goal},\mathbf{u})-\mathbf{p}_g \right\Vert}
\end{equation}
The second term, $d\bar{\mathbf{x}}(t_{goal},\mathbf{u})/d\mathbf{x}(t_{goal},\mathbf{u})$, 
can be computed directly given the  projection function $p$.

All that remains is to compute the third term, $d\mathbf{x}(t_{goal},\mathbf{u})/d\mathbf{u}$.  Given a discrete time dynamics function, this gradient can be computed iteratively via the multivariate chain rule as:
\begin{equation}
    \frac{d\mathbf{x}(t+dt,\mathbf{u})}{d\mathbf{u}} = \frac{\partial\mathbf{x}(t+dt,\mathbf{u})}{\partial\mathbf{u}} + \frac{\partial\mathbf{x}(t+dt,\mathbf{u})}{\partial\mathbf{x}(t,\mathbf{u})} \frac{d\mathbf{x}(t,\mathbf{u})}{d\mathbf{u}}
    \label{eq:pos_deriv}
\end{equation}
We start from $d\mathbf{x}(0,\mathbf{u})/d\mathbf{u} = 0$, as the current position is independent of the upcoming control, and apply Equation~\ref{eq:pos_deriv} iteratively until the gradient at some user-defined time, $T$, is obtained. However, 
we may not have a closed form solution for $\mathbf{x}(t,\mathbf{u})$, and therefore no closed form solution for the discrete time dynamics. 
While we could compute the gradient of the RK4 dynamics directly, as we do for the forward propagation, we find it more computationally efficient (while sufficiently accurate) to compute the gradient using a series of trapezoidal integration steps on the (known) continuous dynamics.

In our case, the trapezoidal integration is defined as follows:
\begin{equation}
\begin{split}
    \mathbf{x}^+ &= \mathbf{x}(t,\mathbf{u})
            +dt\cdot\dot{\mathbf{x}}(\mathbf{x}(t,\mathbf{u}),\mathbf{u}) \\
    \mathbf{x}(t+dt,\mathbf{u}) &\approx \mathbf{x}(t,\mathbf{u})
       + \frac{dt}{2}(\dot{\mathbf{x}}(\mathbf{x}(t,\mathbf{u}),\mathbf{u})
                      + \dot{\mathbf{x}}(\mathbf{x}^+,\mathbf{u}))
\end{split}
\end{equation}
Using these approximate dynamics, we can iteratively compute the partial derivatives required by Equation~\ref{eq:pos_deriv} as follows:
\begin{equation}
    \frac{\partial\mathbf{x}(t+dt,\mathbf{u})}{\partial\mathbf{u}} \approx \frac{dt}{2}\left(\frac{\partial\dot{\mathbf{x}}(\mathbf{x}(t,\mathbf{u}),\mathbf{u})}{\partial\mathbf{u}} + \frac{\partial\dot{\mathbf{x}}(\mathbf{x}^+,\mathbf{u})}{\partial\mathbf{u}}\right)
    \label{eq:trap_u_partial}
\end{equation}
\begin{align}
    \frac{\partial\mathbf{x}(t+dt,\mathbf{u})}{\partial\mathbf{x}(t,\mathbf{u})} & \approx I  +  \dfrac{dt}{2}  \bigg(\frac{\partial\dot{\mathbf{x}}(\mathbf{x}(t,\mathbf{u}),\mathbf{u})}{\partial\mathbf{x}(t,\mathbf{u})}   && \nonumber\\ & \hspace{1.8cm} + \frac{\partial\dot{\mathbf{x}}(\mathbf{x}^+,\mathbf{u})}{\partial\mathbf{x}^+}\frac{\partial\mathbf{x}^+}{\partial\mathbf{x}(t,\mathbf{u})}\bigg)
    \label{eq:trap_x_partial}
\end{align}
where the half step partial derivatives are computed as:
\begin{equation}
    \frac{\partial\mathbf{x}^+}{\partial\mathbf{u}}=dt \frac{\partial\dot{\mathbf{x}}(\mathbf{x}(t,\mathbf{u}),\mathbf{u})}{\partial\mathbf{u}}
\end{equation}
\begin{equation}
    \frac{\partial\mathbf{x}^+}{\partial\mathbf{x}(t,\mathbf{u})} = \mathbf{I} + dt \frac{\partial\dot{\mathbf{x}}(\mathbf{x}(t,\mathbf{u}))}{\partial\mathbf{x}(t,\mathbf{u})}
\end{equation}

The full iterative process for computing the derivative $d\mathbf{x}(T,\mathbf{u})/d\mathbf{u}$ at time $T$ is shown in Algorithm~\ref{alg:pos_control_deriv} (here, we set  $T=t_{goal}$).
Finally, we combine the resulting gradient as in Equation~\ref{eq_dCdU} to compute the total gradient for the goal cost term.

\begin{algorithm}[t]
\SetKwInOut{Input}{Input}
\SetKwInOut{Output}{Output}
\Input{$\mathbf{u},\ \mathbf{x}(0,\mathbf{u}),\ T,\ dt_{max}$}
\Output{$\mathbf{x}(T,\mathbf{u}),\ d\mathbf{x}(T,\mathbf{u})/d\mathbf{u}$}
$t=0$\\
$\dfrac{d\mathbf{x}(t,\mathbf{u})}{d\mathbf{u}}=0$\\

\While{$t < T$}{
    $dt=\min(dt_{max},T-t)$\\
    $\mathbf{x}(t+dt,\mathbf{u})=\mathrm{RK4}(\mathbf{x}(t,\mathbf{u}),\mathbf{u},dt)$\\
    $\dfrac{\partial\mathbf{x}(t+dt,\mathbf{u})}{\partial\mathbf{u}} = \mathrm{Equation~\ref{eq:trap_u_partial}}$\\
    $\dfrac{\partial\mathbf{x}(t+dt,\mathbf{u})}{\partial\mathbf{x}(t,\mathbf{u})} = \mathrm{Equation~\ref{eq:trap_x_partial}}$\\
    $\dfrac{d\mathbf{x}(t+dt,\mathbf{u})}{d\mathbf{u}}= $ \\ \quad \quad \quad \quad $\dfrac{\partial\mathbf{x}(t+dt,\mathbf{u})}{\partial\mathbf{u}} + \dfrac{\partial\mathbf{x}(t+dt,\mathbf{u})}{\partial\mathbf{x}(t,\mathbf{u})} \dfrac{d\mathbf{x}(t,\mathbf{u})}{d\mathbf{u}}$\\
    $t = t+dt$\\
}
\caption{Position Control Gradient}
\label{alg:pos_control_deriv}
\end{algorithm}

\subsubsection{Collision Cost Gradient}
\label{sec:collision_gradient}
Similar to the goal cost term, the collision cost term, $C_{col}$ (Equation~\ref{eq:col_cost}), doesn't depend directly on $\mathbf{u}$, so we must again compute its gradient via the chain rule:
\begin{equation}
    \frac{dC_{col}(\mathbf{u})}{d\mathbf{u}} =
    \frac{dC_{col}(\mathbf{u})}{d\tau(\mathbf{u},\mathbf{o}^*)}\frac{d\tau(\mathbf{u},\mathbf{o}^*)}{d\mathbf{u}}
\end{equation}
where $\mathbf{o}^*$ is the obstacle with the closest time-to-collision.

However, unlike in the goal cost gradient, the time to collision, $\tau$, cannot generally be written explicitly as a function of the controls $\mathbf{u}$,
which prevents us from computing $d\tau(\mathbf{u},\mathbf{o}^*) / d\mathbf{u}$.  
To address this issue, we propose
the use of \emph{implicit differentiation}. This allows us to have an analytic expression of the collision-cost gradient implicitly written as a function of $\tau(\mathbf{u},\mathbf{o}^*)$. 
The implicit relationship between $\tau$ and $\mathbf{u}$ is shown in Equation~\ref{eq:ttc_char}.
Using this relationship, 
we can find $d\tau(\mathbf{u},\mathbf{o}^*) / d\mathbf{u}$
by taking the derivative of Equation~\ref{eq:ttc_char} with respect to $\mathbf{u}_j$ (the $j$th element of the control), and solving for $d\tau(\mathbf{u},\mathbf{o}^*)/d\mathbf{u}_j$:



\begin{equation}
\frac{d\tau(\mathbf{u},\mathbf{o}^*)}{d\mathbf{u}_j} = -\frac{(\bar{\mathbf{x}} - \bar{\mathbf{o}}^*)^T (\dfrac{d\bar{\mathbf{x}}}{d\mathbf{u}_j})}{(\bar{\mathbf{x}} - \bar{\mathbf{o}^*})^T (\dfrac{d\bar{\mathbf{x}}}{d\tau} - \dfrac{d\bar{\mathbf{o}}^*}{d\tau})}
\label{eq:ttc_deriv}
\end{equation}
where $\bar{\mathbf{x}}=\bar{\mathbf{x}}(\tau(\mathbf{u},\mathbf{o}_i),\mathbf{u}))$ and $\bar{\mathbf{o}}^* = \bar{\mathbf{o}}^*(\tau(\mathbf{u},\mathbf{o}^*))$.

Note that $d\bar{\mathbf{x}}/d\tau$ and $d\bar{\mathbf{o}}^*/d\tau$ are the known continuous time dynamics of the robot and the obstacle, and $d\bar{\mathbf{x}}/d\mathbf{u}_j$ can be computed via the chain rule, as shown in section \ref{sec:goal_deriv}, as:
\begin{equation}
\frac{d\bar{\mathbf{x}}}{d\mathbf{u}_j}=\frac{d\bar{\mathbf{x}}}{d\mathbf{x}}\frac{d\mathbf{x}}{d\mathbf{u}_j}
\end{equation}
where $d\bar{\mathbf{x}}/d\mathbf{x}$ is dependent on the dynamics model, and $d\mathbf{x}/d\mathbf{u}_j$ can be computed with Algorithm~\ref{alg:pos_control_deriv}, setting $T=\tau(\mathbf{u},\mathbf{o}_i)$.
If $\tau$ is infinite (i.e. there is no collision), this derivative is 0.

\subsection{State Constraints} \label{sec:state_constr}
When the agent is subject to state constraints, such as an acceleration controlled robot with a maximum velocity constraint, a small modification needs to be applied to NH-TTC.  
Since we are generating trajectories with a single control, we cannot always guarantee that such constraints will be satisfied. For example, applying a non-zero acceleration will eventually violate 
any velocity magnitude constraint.
To address this issue, we enforce state constraints by modifying the continuous time dynamics. 
For example, in the acceleration controlled system, we can provide a (soft) constraint on the velocity to be no more than $v_{max}$ as follows:

\IGNORE{
When the agent is subject to state constraints, such as an acceleration controlled robot with a maximum velocity constraint, a small modification needs to be applied to the above method.  While such state constraints 
could be handled by adding a penalty term in the cost function,  
there is not a natural \icomment{point in time} 
at which to penalize the violation.  \bcomment{As we are optimizing a single control,} if the penalty is applied too soon (temporally), the robot may mistakenly think it can avoid collisions, \iicomment{as the  time to collision search will allow its velocity beyond that time to exceed constraints}.  If the penalty is applied too far in the future, the robot is not allowed to make aggressive accelerations. 
\sjgComment{seems confusing to me} \iicomment{I agree. I think the point is that: one one hand, if we add the penalty too soon, we may overestimate the true time to collision and actually collide. On the other hand, if we add the penalty too late, as it backpropagates, the robot may not be able to take full advantage of its control range.}

To account for this, we enforce these constraints in the continuous time dynamics.  
For example, consider an acceleration controlled system with a velocity constraint.  We can provide a (soft) constraint on the velocity to be no more than $v_{max}$ by modifying
the continuous time dynamics as follows:}

\begin{equation}
    \dot{\mathbf{v}} =
    \begin{cases}
    \dfrac{\mathbf{a}}{100}, &\mathrm{if } \left\Vert\mathbf{v}\right\Vert > v_{max}\ \mathrm{and}\ \mathbf{a}^T\mathbf{v} > 0  \\
    \mathbf{a}, &otherwise
    \end{cases}
\end{equation}
where $\mathbf{a}$ is the acceleration.  While it may be natural to zero out the acceleration if the constraint is violated
this could result in the gradient of the velocity with respect to acceleration going to zero as well, and no optimization would occur. 
Instead, we limit the acceleration to a very small value to avoid the vanishing gradient problem.

The above formulation only enforces the state constraint 
when computing the cost, $C$, and its gradient, $dC/d\mathbf{u}$.
As such, to more strictly enforce the constraint, we also modify the projection in the subgradient descent algorithm (Algorithm~\ref{alg:sgd}).  In addition to the projection onto the control constraint set, we also project the control such that, one timestep in the future, the state constraint isn't violated.  
In the above acceleration controlled system, for example, by solving for the new velocity, we can modify the acceleration to enforce the state constraint as follows:
\begin{equation}
\mathbf{a}_{proj} = 
\begin{cases}
\mathbf{a} &\mathrm{if } \left\Vert \mathbf{v} + dt\cdot\mathbf{a} \right\Vert \leq v_{max} \\
\dfrac{\mathbf{v}^* - \mathbf{v}}{dt}, &otherwise
\end{cases}
\end{equation}
where $\mathbf{v}^*$ is $\mathbf{v} + dt\cdot\mathbf{a}$ projected onto $v_{max}$.
Note that this method is only applicable when the state constraints are on states we can solve for explicitly.

\section{Dynamics Models} \label{sec:dynamics_models}

The exact form of the cost gradient computation (Equations ~\ref{eq_dCdU}-\ref{eq:ttc_deriv}) will vary based on the dynamics of the robot under consideration. 
For the sake of illustration, we show in Section~\ref{sec_case_study} how to apply our method to a robot with a smooth car dynamics model inspired by an autonomous driving task. 
As this smooth car model is relatively complex, with 2nd order, acceleration-controlled, car-like dynamics, we step through all the relevant equations needed to apply our framework in detail. Our work supports a large variety of different types of dynamics models, a few of which are discussed more briefly in Section~\ref{sec_other_dynamics}.


\subsection{Case Study: Smooth Car} \label{sec_case_study}

\begin{figure}
    \centering
    \includegraphics[width=0.5\columnwidth]{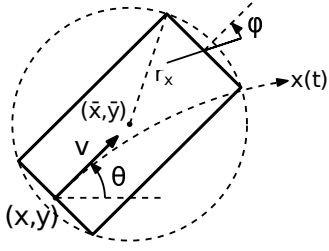}
    \caption{\textbf{Smooth Car}: Definitions of the states of the smooth car $(x,y,\theta,v,\phi)$, in addition to the collision disk $(\bar{x},\bar{y},r_x)$.  Given zero controls, the car will follow the trajectory labeled $\mathbf{x}(t)$.
    }
    \label{fig:car_draw}
\end{figure}

We define a smooth car by a 5d state space in which each state is represented by 2d position, orientation, linear velocity, and steering angle as $\mathbf{x}=(x,y,\theta,v,\phi)$. See Figure~\ref{fig:car_draw} for a visual. 
To make the velocity and the steering angle of the car vary continuously over time, we define a 2d control $\mathbf{u}=(a,\psi)$ that represents the car's linear acceleration and the rate of change of the steering angle. 
In addition to constraints on $a$ and $\psi$, we also impose state constraints on $v$ and $\phi$. 
As described in section~\ref{sec:state_constr}, we introduce the following functions to help enforce the state constraints:
\begin{equation}
\begin{aligned}
    k_a(v,a) &= 
    \begin{cases}
    1 &\mathrm{if}\ \left\vert v \right\vert > v_{max}\ \mathrm{and}\ a\cdot v > 0 \\
    \frac{1}{100} &otherwise
    \end{cases} \\
    k_\psi(\phi,\psi) &= 
    \begin{cases}
    1 &\mathrm{if}\ \left\vert \phi \right\vert > \phi_{max}\ \mathrm{and}\ \psi\cdot\phi > 0 \\
    \frac{1}{100} &otherwise
    \end{cases}
\end{aligned}
\end{equation}
Using these functions, we can define the continuous time dynamics of the system as:
\begin{equation}
    \begin{aligned}
    \dot{x}&=v\cos(\theta) & \dot{y}&=v\sin(\theta) &
    \dot{\theta}&=v\tan(\phi)/L \\
    \dot{v}&= k_a(v,a)\ a & \dot{\phi}&= k_\psi(\phi,\psi)\ \psi
    \end{aligned}
    \label{eq:scar_cont_dyn}
\end{equation}
where $L$ is the length of the car. Note that $\dot{v}$ and $\dot{\phi}$ are modified to maintain the velocity and steering angle constraints during forward propagation.

From these dynamics, we can compute the partial derivatives with respect to both the state and the controls (only non-zero derivatives are shown).  The partial derivatives with respect to the state are:
\begin{equation}
\begin{aligned}
\frac{\partial\dot{x}}{\partial \theta}&=-v\sin(\theta) &
\frac{\partial\dot{x}}{\partial v}&=\cos(\theta) \\
\frac{\partial\dot{y}}{\partial \theta}&=v\cos(\theta) &
\frac{\partial\dot{y}}{\partial v}&=\sin(\theta) \\
\frac{\partial\dot{\theta}}{\partial v}&=\tan(\phi)/L &
\frac{\partial\dot{\theta}}{\partial \phi}&=v/(L\cos^2(\phi))
\end{aligned}
\end{equation}
and the partial derivatives with respect to the controls are:
\begin{equation}
\begin{aligned}
\frac{\partial\dot{v}}{\partial a}&= k_a(v,a) &
\frac{\partial\dot{\phi}}{\partial \psi}&= k_\psi(\phi,\psi)
\end{aligned}
\end{equation}

Using the continuous time dynamics and these partial derivatives, we can compute discrete time dynamics via trapezoid integration, and find the partial derivatives necessary for Algorithm~\ref{alg:pos_control_deriv}.  The discrete time dynamics are approximated as:
\begin{equation}
    \begin{aligned}
    x_{t+dt}&\approx x_t + \frac{dt}{2} (v_t \cos(\theta_t) + v_{t+dt} \cos(\theta_{t+dt})) \\
    y_{t+dt}&\approx x_t + \frac{dt}{2} (v_t \sin(\theta_t) + v_{t+dt} \sin(\theta_{t+dt}))\\
    \theta_{t+dt}&\approx \theta_t + \frac{dt}{2L} (v_t\tan(\phi_t) + v_{t+dt}\tan(\phi_{t+dt})) \\
    v_{t+dt}&\approx v_t + dt\ a\ k_a(v_t,a) \\
    \phi_{t+dt}&\approx \phi_t + dt\ \psi\ k_\psi(\phi_t,\psi)
    \end{aligned}
\end{equation}
Using this, we can analytically compute the form of the partial derivatives with respect to the state at time $t$ and the controls, as required by Equation~\ref{eq:pos_deriv} (again only showing non-zero elements). 
First, the partials of the $x$-component of the state with respect to the controls are:
\begin{equation}
    \begin{aligned}
    \frac{\partial x_{t+dt}}{\partial x_t} &= 1 \\
    \frac{\partial x_{t+dt}}{\partial \theta_t} &= -\frac{dt}{2}(v_t\sin(\theta_t)+v_{t+1}\sin(\theta_{t+1})) \\
    \frac{\partial x_{t+dt}}{\partial v_t} &= \frac{dt}{2}(\cos(\theta_t)+\cos(\theta_{t+1})-v_{t+1}\sin(\theta_{t+1})\frac{\partial\theta_{t+1}}{\partial v_t}) \\
    \frac{\partial x_{t+dt}}{\partial \phi_t} &= -\frac{dt}{2}v_{t+1}\sin(\theta_{t+1})\frac{\partial\theta_{t+1}}{\partial \phi_t} \\
    \frac{\partial x_{t+dt}}{\partial a} &= \frac{dt}{2}(dt\cos(\theta_{t+dt}) - v_{t+1}\sin(\theta_{t+1})\frac{\partial \theta_{t+1}}{\partial a}) \\
    \frac{\partial x_{t+dt}}{\partial \psi} &= -\frac{dt}{2}v_{t+1}\sin(\theta_{t+dt})\frac{\partial \theta_{t+1}}{\partial \psi} \\
    \end{aligned}
    \label{eq:scar_dx_dx}
\end{equation}
Similarly, the partials of the $y$-component of the state are:
\begin{equation}
    \begin{aligned}
    \frac{\partial y_{t+dt}}{\partial x_t} &= 1 \\
    \frac{\partial y_{t+dt}}{\partial \theta_t} &= \frac{dt}{2}(v_t\cos(\theta_t)+v_{t+1}\cos(\theta_{t+1})) \\
    \frac{\partial y_{t+dt}}{\partial v_t} &= \frac{dt}{2}(\sin(\theta_t)+\sin(\theta_{t+1})+v_{t+1}\cos(\theta_{t+1})\frac{\partial\theta_{t+1}}{\partial v_t}) \\
    \frac{\partial y_{t+dt}}{\partial \phi_t} &= \frac{dt}{2}v_{t+1}\cos(\theta_{t+1})\frac{\partial\theta_{t+1}}{\partial \phi_t} \\
    \frac{\partial y_{t+dt}}{\partial a} &= \frac{dt}{2}(dt\sin(\theta_{t+dt}) + v_{t+1}\cos(\theta_{t+1})\frac{\partial \theta_{t+1}}{\partial a}) \\
    \frac{\partial y_{t+dt}}{\partial \psi} &= \frac{dt}{2}v_{t+1}\cos(\theta_{t+dt})\frac{\partial \theta_{t+1}}{\partial \psi} \\
    \end{aligned}
\end{equation}
The partials of the orientation $\theta_{t+1}$ are:
\begin{equation}
    \begin{aligned}
    \frac{\partial\theta_{t+1}}{\partial\theta_t} &= 1 && \\
    \frac{\partial\theta_{t+1}}{\partial v_t} &= \frac{dt}{2L}(\tan(\phi_t)+\tan(\phi_{t+1})) \\
    \frac{\partial\theta_{t+1}}{\partial \phi_t} &= \frac{dt}{2L}(\frac{v_t}{\cos^2(\phi_t)}+\frac{v_{t+1}}{\cos^2(\phi_{t+1})}) \\
    \frac{\partial\theta_{t+1}}{\partial a} &= \frac{dt^2\,k_a(v_t,a)}{2L}\tan(\phi_{t+1}) \\
    \frac{\partial\theta_{t+1}}{\partial \psi} &= \frac{dt^2\,v_{t+1}\,k_\psi(\phi_t,\psi)}{2L\cos^2(\phi_{t+1})} 
    \end{aligned}
\end{equation}
Finally the partials with respect to the velocity $v$ and steering angle $\phi$ are:
\begin{equation}
    \begin{aligned}
    \frac{\partial v_{t+1}}{\partial v_t} &= 1 \\
    \frac{\partial v_{t+1}}{\partial a} &= dt\,k_a(v_t,a) \\
    \end{aligned}
\end{equation}
and:
\begin{equation}
    \begin{aligned}
    \frac{\partial \phi_{t+1}}{\partial \phi_t} &= 1 \\
    \frac{\partial \phi_{t+1}}{\partial \psi} &= dt\,k_\psi(\phi_t,\psi)
    \end{aligned}
    \label{eq:scar_trap_ddyn}
\end{equation}

We also need to define the function, $p$, mapping the state, $\mathbf{x}$, to the Euclidean workspace, and its derivatives.  Because we are modeling the car from the center of the rear axle, we can minimize the encompassing area of the collision avoidance circle by shifting the collision center to lie on the center of the car rather than on the real axle (Figure~\ref{fig:car_draw}):
\begin{equation}
    \begin{aligned}
    \bar{x} &= x + \frac{L}{2}\cos(\theta) \\
    \bar{y} &= x + \frac{L}{2}\sin(\theta) 
    \end{aligned}
    \label{eq:xbar_def}
\end{equation}
The non-zero derivatives of the collision disk center are:
\begin{equation}
    \begin{aligned}
    \frac{\partial\bar{x}}{\partial x} &= 1 &
    \frac{\partial\bar{y}}{\partial y} &= 1 \\
    \frac{\partial\bar{x}}{\partial \theta} &= -\frac{L}{2}\sin(\theta) &
    \frac{\partial\bar{y}}{\partial \theta} &= \frac{L}{2}\cos(\theta)
    \end{aligned}
    \label{eq:scar_dxbar_dx}
\end{equation}

The above derivatives, along with those of the continuous dynamics (Equation~\ref{eq:scar_cont_dyn}), and those of the trapezoidal integration (Equations~\ref{eq:scar_dx_dx}-\ref{eq:scar_trap_ddyn}), fully define the goal cost gradient (Equation~\ref{eq_dCdU}).

To compute collisions, we also define the radius of the collision disk, assuming a 2-to-1 length to width ratio for the car:
\begin{equation}
    r_x = \frac{L\sqrt{5}}{4}
    \label{eqn_carradius}
\end{equation}
Equation~\ref{eqn_carradius} together with the continuous dynamics (Equation~\ref{eq:scar_cont_dyn}) and the offset collision circle center (Equation~\ref{eq:xbar_def}), allows us to compute the linear continuous collision checks between RK4 integration steps in order to estimate the time to collision with any obstacles in the scene.  After computing the time to collision, we can compute the collision cost, using  Equation~\ref{eq:col_cost}, and the collision cost gradient, by combining Equations~\ref{eq:ttc_deriv}, \ref{eq:scar_dx_dx}-\ref{eq:scar_trap_ddyn}, and~\ref{eq:scar_dxbar_dx}.  Combining the collision cost gradient with the goal cost gradient gives the full gradient, which can then be used in Algorithm~\ref{alg:sgd} for the control update.

\subsection{Other Dynamics Models} \label{sec_other_dynamics}
While our framework supports many robot dynamics models, for the majority of our results, we consider the following five different models that span a range of 1st and 2nd order dynamics, with or without kinematic constraints, and includes both holonomic and non-holonomic systems.

\begin{itemize}
    \item Velocity (V):  Here the state is the 2d position ($x,y$), and the controls are the 2d velocities ($v_x,v_y$).  The continuous time dynamics are:
    \begin{equation}
    \begin{aligned}
    \dot{x}&=v_x &  \dot{y}&=v_y
    \end{aligned}
    \end{equation}
    \item Acceleration (A):  Here the state is the 2d position and the 2d velocity ($x,y,v_x,v_y$), and the controls are the 2d accelerations ($a_x,a_y$).  The continuous time dynamics are:
    \begin{equation}
    \begin{aligned}
    \dot{x}&=v_x & \dot{y}&=v_y &
    \dot{v}_x&=a_x & \dot{v}_y&=a_y
    \end{aligned}
    \end{equation}
    \item Differential Drive (DD):  Here the state is the 2d position and the 1d orientation ($x,y,\theta$), and the controls are the linear and angular velocities ($v,\omega$).  The continuous time dynamics are:
    \begin{equation}
    \begin{aligned}
    \dot{x}&=v\cos(\theta) &
    \dot{y}&=v\sin(\theta) &
    \dot{\theta}&=\omega
    \end{aligned}
    \end{equation}
    \item Smooth Differential Drive (SDD):  Here the state is the 2d position, the 1d orientation, and the linear and angular velocity ($x,y,\theta,v,\omega$), and the controls are the linear and angular accelerations ($a,\alpha$).  The continuous time dynamics are:
    \begin{equation}
    \begin{aligned}
    \dot{x}&=v\cos(\theta) &
    \dot{y}&=v\sin(\theta) &
    \dot{\theta}&=\omega \\
    \dot{v}&=a &
    \dot{\omega}&=\alpha
    \end{aligned}
    \label{eq:sdd}
    \end{equation}
    \item Simple Car (Car):  Here the state is the 2d position and the 1d orientation ($x,y,\theta$), and the controls are the linear velocity and the steering angle ($v,\phi$).  The continuous time dynamics are:
    \begin{equation}
    \begin{aligned}
    \dot{x}&=v\cos{\theta} &
    \dot{y}&=v\sin{\theta} &
    \dot{\theta}&=v\tan{\phi}/L
    \end{aligned}
    \end{equation}
    where $L$ is the length of the car.
\end{itemize}

\section{Implementation Details}
\label{sec:implementation}
In all of our experiments, unless otherwise specified, we use the following constraints for the linear velocity,  angular velocity, linear acceleration, angular acceleration, steering angle, and steering angle velocity, respectively: 
$v\,=\,0.3\,\mathrm{m/s}$, $\omega\,=\,1.0\, \mathrm{rad/s}$, $a\,=\,1.0\,\mathrm{m/s^2}$, $\alpha\,=\,\pi\,\mathrm{rad/s^2}$, $\phi\,=\,\pi/4\,\mathrm{rad}$, and $\psi\,=\,\pi/4\,\mathrm{rad/s}$.
The cost parameters, $\kappa_{goal}$ and $\kappa_{col}$, are both set to 1.  For the time-to-collision search, $t_{horiz}$ is set to 5$\,$s and $dt_{max}$ is set to 0.1$\,$s.  The time when we compute the goal distance, $t_{goal}$, is set to 1$\,$s.  Trajectories are planned using 10$\,$ms of planning time, and controls are updated at 10$\,$Hz.
All results generated on a single thread on a Intel Xeon 3.0GHz processor.  For the real robot results, each agent planned in its own thread.  We implemented our subgradient-based optimization framework in C++, using Eigen~\citep{eigenweb} to efficiently handle matrix and vector operations.

Our implementation contains a few algorithmic optimizations.  First, as the obstacle trajectories are static throughout all optimization iterations per planning step, we pre-compute the trajectories at the beginning of each planning step.  Second,
we compute the collision checks for each linear segment against every obstacle before moving to the next linear segment.  This allows the time-to-collision search to exit as soon as the first collision is found.  As such, collision times can be quick to compute for nearby obstacles, allowing for many optimization iterations even in dense scenarios.  Finally, we chose a high resolution for $dt_{max}$, such that our time-to-collision search was very accurate for every dynamics model. However, for dynamics where RK4 and the linear collision check are sufficiently accurate over longer time periods (such as an acceleration controlled system), $dt_{max}$ could be increased to greatly improve performance.  This is due to the fact that the max number of iterations of the collision search is defined by $t_{horiz}/dt_{max}$.

In the following sections, the figures are rendered as follows:
all agents are colored by their dynamics model, or grey if they are non-reactive to the controlled robots.  Velocity agents are colored dark blue, acceleration agents are green, differential drive agents are red, smooth differential drive agents are cyan, simple car agents are yellow, and smooth car agents are orange.  For models with orientation in their state, a black arrow shows their forward direction.  Simple car and smooth car agents are drawn with both their physical extents and the corresponding collision disks.  Goal positions are marked by X marks, with the color corresponding to the dynamics model of the agent.  For a video of all robot experiments and planned trajectories, please see Extension~1.

\section{Single-Agent Collision Avoidance} \label{sec:collision_avoidance}

A key application of anticipatory collision avoidance is to allow a robot to avoid nearby dynamic obstacles as it moves to its goal. 
Below, we consider two scenarios in simulation, where the agent is navigating in environments containing non-reactive obstacles.

\subsection{Three Obstacle Scenario}
\begin{figure}
    \centering
    \captionsetup[subfigure]{justification=centering}
    \null\hfill
    \subfloat[Velocity Robot]{
    \fbox{\includegraphics[width=0.6\columnwidth]{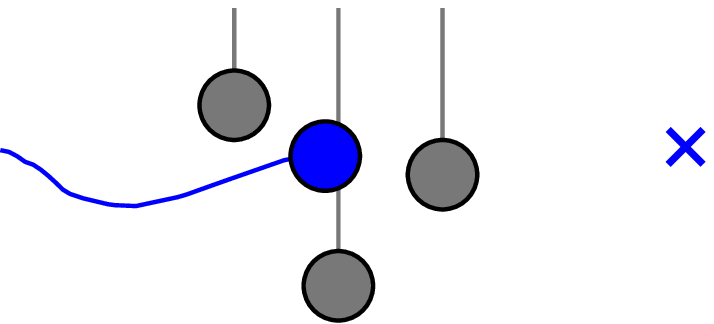}}
    }
    \hfill\null \\
    \null\hfill
    \subfloat[Acceleration Robot]{
    \fbox{\includegraphics[width=0.6\columnwidth]{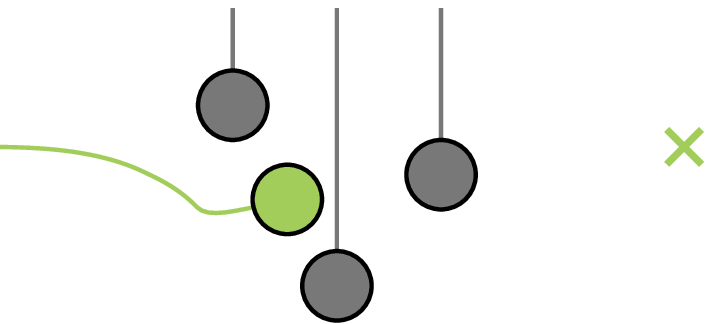}}
    }
    \hfill\null
    \caption{\textbf{Three-Obstacles}: A single robot avoids three non-reactive linear velocity obstacles, while navigating from left to right. In (a), the robot is velocity controlled.  In (b), it is acceleration controlled. 
    The trajectories traced by the robot and the obstacles so far are displayed as solid lines.
    The X marks denote the goal location of each robot. 
    }
    \label{fig:3v1}
\end{figure}

In this scenario, a single agent avoids 3 non-reactive linear velocity obstacles, each with a different velocity, while traversing from left to right.  We show results in Figure~\ref{fig:3v1} for both a velocity controlled agent and an acceleration controlled agent.  Both dynamics models are able to find paths that slip between the first and second obstacle. Note that changing dynamics models from velocity-controlled to acceleration-controlled has a noticeable effect on the robot's path, with the acceleration-controlled robot taking a smoother path due to the bounds on the allowed acceleration. Both robots reach their goals at about the same time (8\,s for the acceleration controlled vs 8.1\,s for the velocity controlled). This is because the acceleration-controlled robot still has a limit on the velocity it can take implemented as a state constraint.

\begin{figure*}
\centering
\captionsetup[subfigure]{justification=centering}
\null\hfill
\subfloat[$t=8.5\,$s]
{\fbox{\includegraphics[width=0.31\textwidth]{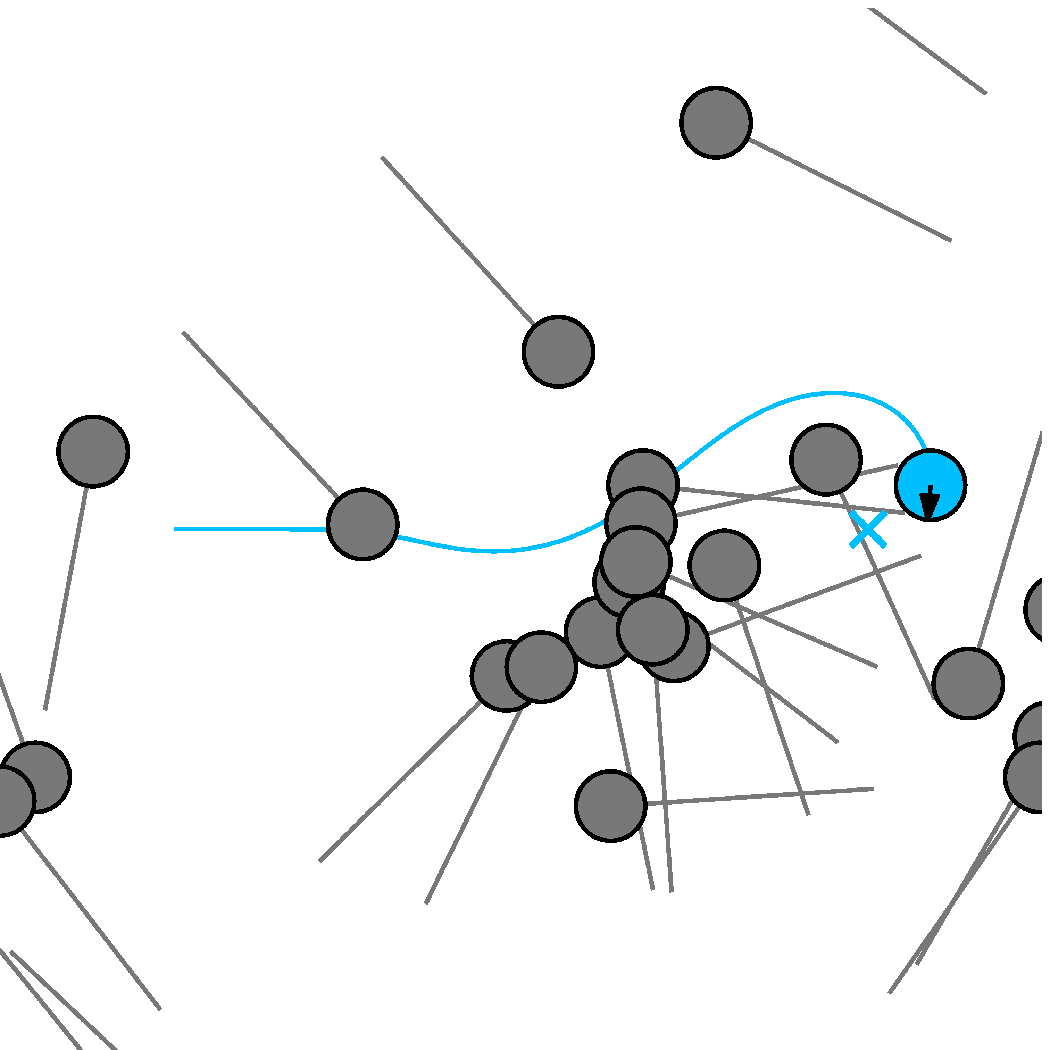}}} 
\hfill
\subfloat[$t=13.2\,$s]
{\fbox{\includegraphics[width=0.31\textwidth]{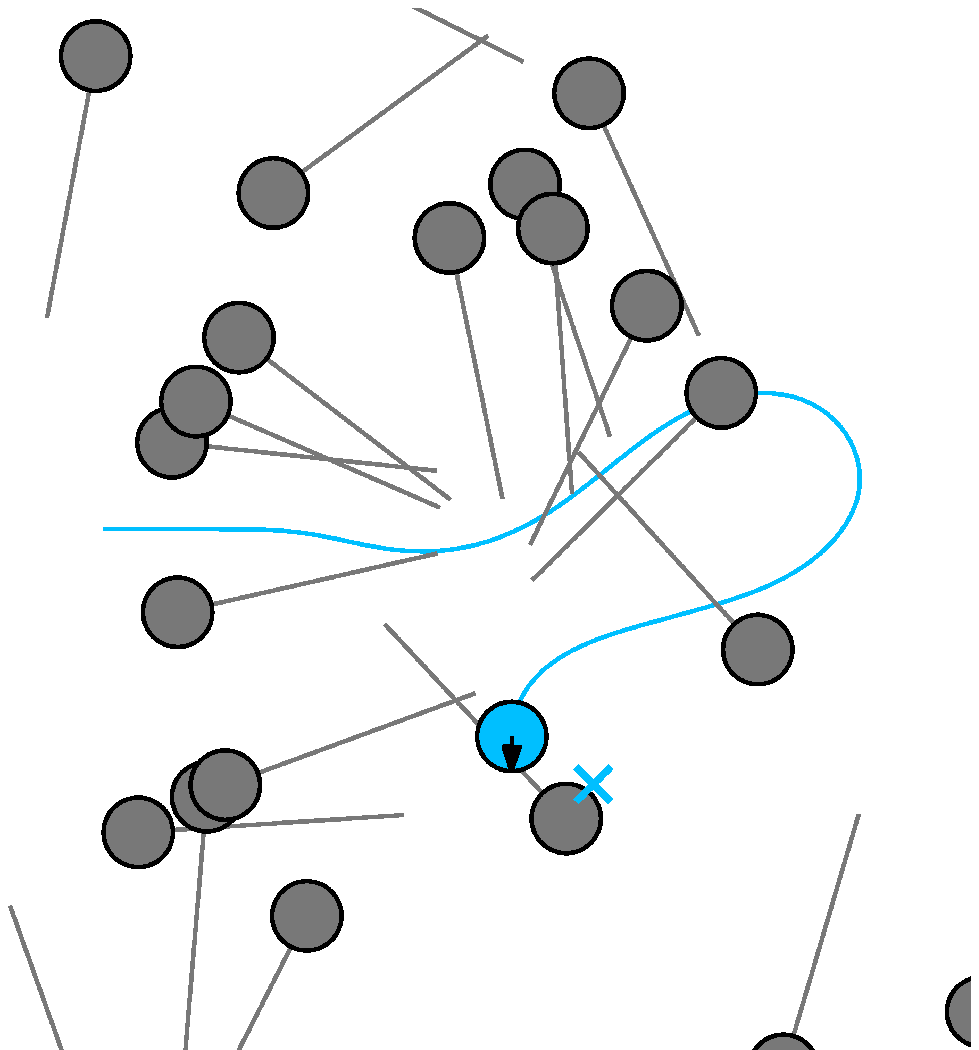}}}
\hfill
\subfloat[$t=26.\,$s]
{\fbox{\includegraphics[width=0.31\textwidth]{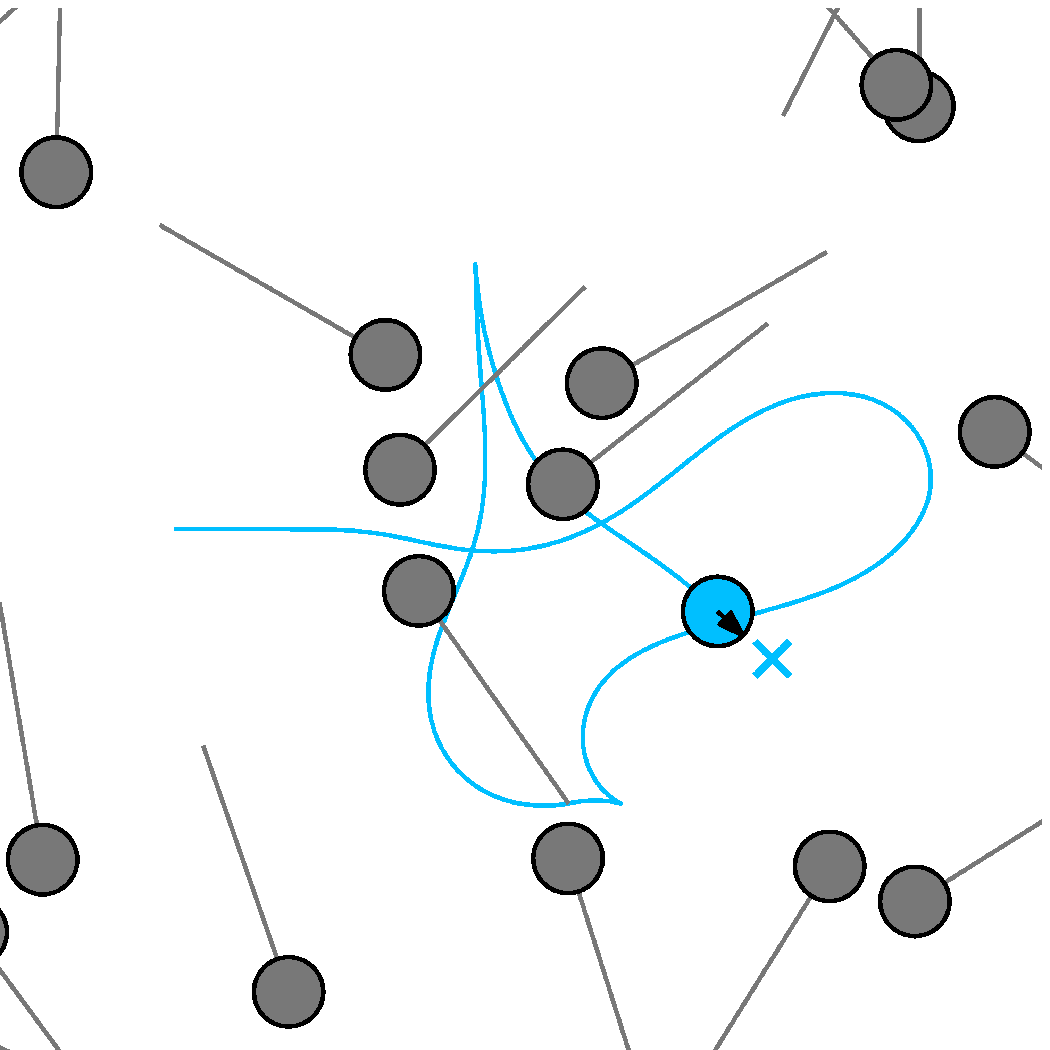}}}
\hfill\null
\caption{\textbf{Smooth Differential Drive Robot Among Random Obstacles}:
The cyan robot moves to a random goal position, indicated by the colored x, while avoiding a number of grey dynamic obstacles following linear paths. 
The recent trajectories traced by the robot and the obstacles are displayed with solid lines. 
}
    \label{fig:random_v}
\end{figure*}

\subsection{Random Velocity-Agents Scenario} \label{sec:random_obsts}
In a more challenging scenario, an agent is attempting to navigate to randomly generated goals  while avoiding 40 non-reactive linear velocity obstacles. 
Time lapses from one trial of this scenario are shown in Figure~\ref{fig:random_v}.
This is a rather difficult scenario as the randomly moving agents can box the agent in, and create scenarios that are impossible to avoid collisions in. 
Even though such challenging situations are unlikely to happen in real life, they allow us to experimentally evaluate the  performance of our method in extreme scenarios. Note, this is the only scenario we tested for which there are any collisions with our method.

Table~\ref{tab:col_frames} shows how our method performs in this scenario with different agent dynamics models.  In general, increasing the complexity of the dynamics model increases the average number of colliding frames.  Overall though, even in this very challenging scenario, collisions are very rare, typically occurring in less than 0.5\% of frames.
Even in the worst average case, where a smooth car-like robot has only $1\,$ms of optimization time, collisions occur in only 1.3\% of the frames.
For nearly all of the dynamics models, increasing the optimization time both reduces the average number of colliding frames and the variance in the number of colliding frames.  
It is important to note, though, that most of the cost improvement occurs within the first 5$\,$ms.

\begin{table}[t]
\centering
\begin{tabular}{c|ccc}
     & \multicolumn{3}{c}{Optimization Time} \\
    Dynamics & 1\,ms & 5\,ms & 10\,ms \\
     \hline
     V & $99.7\%\ (1.4)$ & $99.9\%\ (0.4)$ & $99.9\%\ (0.7)$ \\
     A & $99.7\%\ (1.1)$ & $99.9\%\ (0.6)$ & $99.9\%\ (0.6)$ \\
     DD & $99.5\%\ (1.4)$ & $99.5\%\ (1.4)$ & $99.6\%\ (1.1)$ \\
     SDD & $99.0\%\ (1.8)$ & $99.4\%\ (1.5)$ & $99.5\%\  (1.3)$ \\
     Car & $99.0\%\ (1.9)$ & $99.5\%\ (1.2)$ & $99.6\%\ (1.1)$ \\
     SCar & $98.7\%\ (2.2)$ & $99.6\%\ (1.1)$ & $99.7\%\ (0.9)$ \\
     \vspace{0.2pt}
\end{tabular}
\caption{\textbf{Percent of Collision-Free Frames}: The average percent and standard deviation (reported in percentage points) of collision-free frames is shown for the scenario in Figure~\ref{fig:random_v} for various dynamics models. Experiments were run for 1000 frames, and averaged over 1000 runs per dynamics model.
}
\label{tab:col_frames}
\end{table}

\section{Extending NH-TTC} \label{sec:extensions}
The NH-TTC framework we propose can easily be extended to tasks beyond a single robot navigating around dynamic obstacles. In Section~\ref{sec:dyn_goals} we show how to adapt the cost function to support a robot chasing a dynamic target. In Section~\ref{sec:reciprocity} we show how to extend the framework to support decentralized navigation of multiple controlled robots in a shared environment.

\subsection{Dynamic Goals} 
\label{sec:dyn_goals}
\begin{figure}
\centering
\null\hfill
\includegraphics[width=0.95\columnwidth]{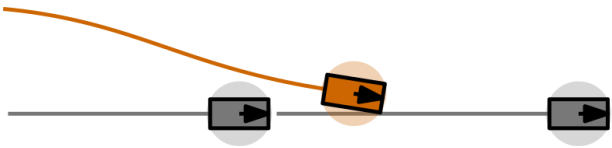}
\hfill\null
\caption{\textbf{Car Following}: A smooth car-like robot, shown in orange, navigates to follow the lead car on the right.}
    \label{fig:overtake}
\end{figure}

The goal cost function defined in Equation~\ref{eq:goal_cost} focuses on a single static goal. However, in many cases robots can have dynamic goals (e.g., chasing a moving target). In such cases, the robot needs to understand the moving nature of its goal to successfully reach it. Our 
framework can be easily adapted to support such dynamic goals. 
Simply greedily minimizing the distance to a dynamic goal can result in oscillations around the goal as it moves.  To remedy this oscillatory behavior, we can compute the goal cost, $\mathcal{C}_{goal}$, at multiple temporal points, and then average those costs: 
\begin{equation}
    \mathcal{C}_{dyn\_goal}=\frac{1}{n}\sum_{i=1}^{n}C_{goal}(\mathbf{u},\mathbf{p}_{goal}^{t_{\,i}})
\end{equation}
This requires the robot not only to reach the goal, but then to stay on top of it as it keeps moving.  In effect, it requires the robot's velocity to synchronize with that of the goal.  We show this type of dynamic goal behavior in the following scenario:

\paragraph{Car-following.} 
In this scenario, a smooth car-like robot is attempting to follow a passive car moving at a constant speed, while a third slower moving car has gotten between them. The robot tries to maintain a small following distance behind the lead obstacle, but the trailing obstacle starts too close for the robot to move in between. As such, the car needs to wait for a gap to open between the obstacles before sliding in. See Figure~\ref{fig:overtake} for an overview.

\subsection{Reciprocity in Multirobot Scenarios} \label{sec:reciprocity}
\begin{figure}
\centering
\captionsetup[subfigure]{justification=centering}
\subfloat[No Reciprocity]
{\includegraphics[width=0.75\columnwidth]{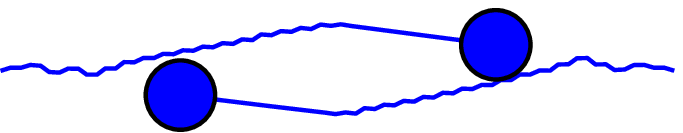}} \\ 
\subfloat[With Reciprocity]
{\includegraphics[width=0.75\columnwidth]{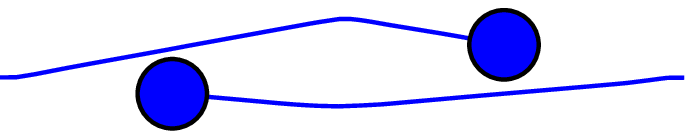}}
\caption{\textbf{Reciprocity}: 
Variants of a two agent scenario obtained with our framework, where the velocity controlled agents want to swap sides.  In (a), reciprocity is disabled and each agent takes full responsibility to resolve the collision.  In (b), reciprocity is enabled using Equation~\ref{eq:reciprocity}. 
This shows the smoothing effect reciprocity has on the resulting trajectories.
}
    \label{fig:reciprocity}
\end{figure}

Simply optimizing the cost function in equation~\ref{eq:cost_func} is not the correct behavior to take when the obstacles are also actively avoiding collisions.  In these scenarios, avoiding the full collision at every time step can result in oscillatory behavior.  
This is due the fact that each agent tries to resolve the collision by itself without accounting for the fact that the other agent, by symmetry, is facing the exact same condition. Consequently, once each agent chooses a control that resolves the collision, it will then select a goal-directed control which will introduce a new collision that needs to be resolved. This pattern of alternating controls will continue until the two agents move past each other leading to jerky motion (see Figure~\ref{fig:reciprocity}a).

A common approach to address this issue is to 
enable \emph{reciprocity} between the two agents which allows the agents to share the effort of averting the collision during mutual collision avoidance tasks~\citep{rvo}. In particular,
inspired by the approach taken by ORCA (only avoiding half the collision), we only update the control to halfway between the new optimal control and the previous control:
\begin{equation}
    \mathbf{u}_{new} = \frac{1}{2}(\mathbf{u}_{prev} + \mathbf{u}^*)
    \label{eq:reciprocity}
\end{equation}
This modification results in significantly smoother paths, while still fully avoiding collisions (see Figure~\ref{fig:reciprocity}b).  Even if the obstacle is not avoiding the agent, the agent will still converge to a collision-free control after a few time steps.

In the following set of results, there are multiple agents planning in a  decentralized manner.

\subsection{2 vs 1 Oncoming}

\begin{figure}
\captionsetup[subfigure]{justification=centering}
\centering
\subfloat[Velocity Robots (V) - ORCA (\cite{orca})]
{\includegraphics[width=0.7\columnwidth]{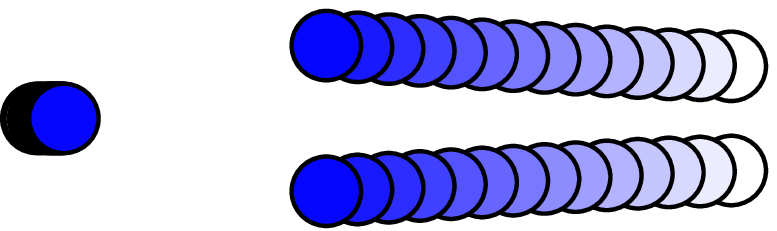}} 
\\
\subfloat[Velocity Robots (V) - NH-TTC]
{\includegraphics[width=0.7\columnwidth]{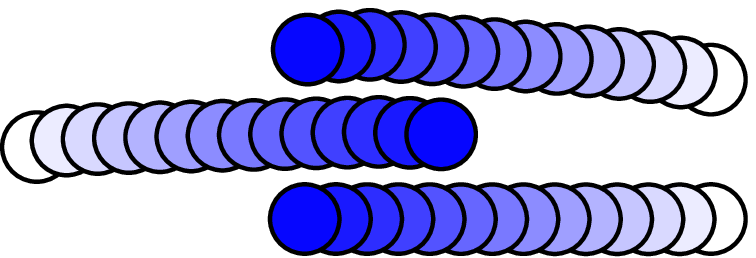}} 
\\
\subfloat[Differential Drive Robots (DD) - NH-TTC]
{\includegraphics[width=0.7\columnwidth]{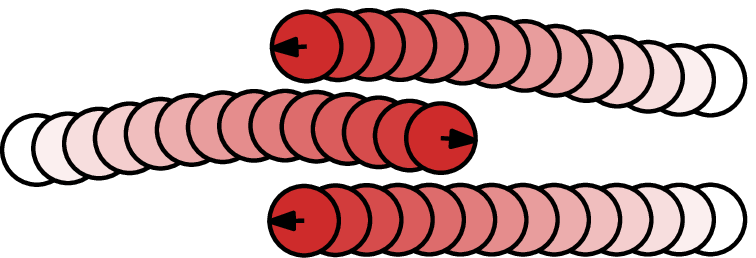}}
\caption{\textbf{2 vs 1 Oncoming}: 
Two simulated robots move from right to left while a third moves from left to right. All figures show the robots after 4 seconds of simulation. The traces of the robots  are shown as colored disks which are
light at their initial positions and dark at their current positions. 
(a) Using the ORCA framework, the standalone robot is reluctant to move forward until the other two robots have walked around it. (b) In contrast, using NH-TTC, all three robots are able to safely reach to their goals in a timely manner. (c) Similar behavior is obtained when NH-TTC plans for differential-drive agents.
 }
\label{fig:2v1}
\end{figure}

In this scenario, two agents standing on one side of an environment have to move toward a third agent that is standing on the opposite side. 
As shown in Figure~\ref{fig:2v1}a, using the ORCA framework to plan for holonomic, velocity-controlled agents results in the lone agent staying put until the other two agents have moved around it. 
This is due to the fact that ORCA conservatively approximates the set of forbidden velocities with  half-planes throwing away too many feasible velocities that the agents could have taken. In contrast, by using our subgradient-based optimization framework, all three agents are able to quickly resolve the collisions and reach to their goals in a timely manner as depicted in Figure~\ref{fig:2v1}b. In addition, as compared to ORCA and many of its
extensions, such as \cite{mora1} and \cite{gorca}, 
our approach plans directly in the agent's control space allowing us to find smooth and collision-free paths for different motion models without being required to cast controls into an intermediate velocity space. As an example, see Figure~\ref{fig:2v1}c for  trajectories obtained by NH-TTC for differential-drive robots.

\subsection{Heterogeneous Circle}
\begin{figure}
\centering
\fbox{\includegraphics[width=0.65\columnwidth]{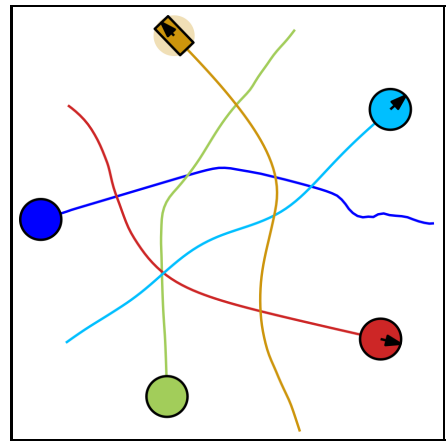}}
\caption{\textbf{5-Agent Circle}: Each of the agents attempts to move to its antipodal position on a circle.  The dark blue agent is velocity controlled, the green agent is acceleration controlled, the red agent is differential drive controlled, the light blue agent is smooth differential drive controlled, and the yellow agent is a simple car.}
    \label{fig:circle_5}
\end{figure}

To highlight how our approach can handle interactions between heterogeneous agents, i.e. agents that can have different motion models and state spaces, we consider a scenario with five agents, each with a different 
robot model highlighted in Section~\ref{sec_other_dynamics}. 
The agents are attempting to move to antipodal points on a circle. Figure~\ref{fig:circle_5} shows the paths taken by each agent. As it can also be observed in the companion video,
NH-TTC generates controls that lead to collision-free and smooth paths.  Note that the jitter in the initial portion of the velocity-controlled agent comes from the implicit coordination between the agents. After the first few seconds, once the agents have come to an implicit consensus on the paths to take, the paths are smooth for the rest of execution.


\section{Analysis and Experimentation}\label{sec:analysis_experiments}
\subsection{Physical Robot Results}
To test the applicability of our method to real robots, we implemented our framework on three Turtlebot2 robots (a differential drive system).  We used an OptiTrack system for position and orientation localization, and used the internal odometry of the robots to get the linear and angular velocities. Each robot communicated its current pose and velocity to the other robots, and asynchronously planned over the latest received state of the world. 

\begin{figure}
\captionsetup[subfigure]{justification=centering}
\centering
\subfloat[Following]
{\includegraphics[width=0.7\columnwidth]{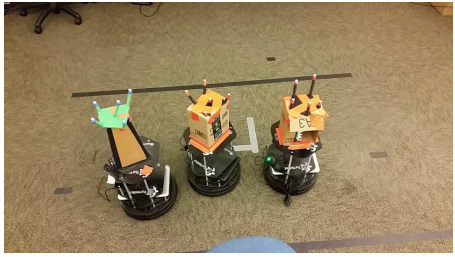}}
\quad
\vspace{0.2pt}
\subfloat[Robot Controls]
{\includegraphics[width=0.8\columnwidth]{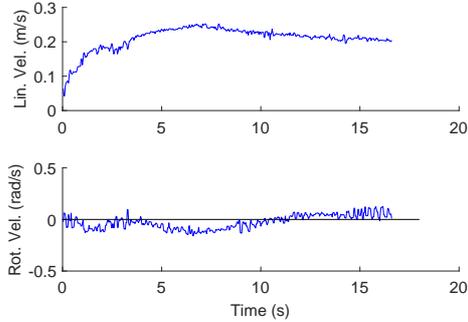}}
\caption{\textbf{Physical Robots - Following Scenario}: One robot attempts to slip between two unresponsive robots (a simulated version is shown in Figure~\ref{fig:overtake}). In (b), the controls executed by the controlled robot are shown, as measured by wheel encoders and gyroscope.
The robot is able to quickly accelerate and slide between the two agents as soon as an appropriate gap opens up. 
}

\label{fig:real_robot_lane}
\end{figure}

We tested the applicability of our approach to physical robots on two scenarios:
\begin{itemize}
\item\emph{``Car" Following}:
Two non-reactive robots moving at a constant velocity of 0.2 and 0.15$\,$m/s, respectively, with a single controlled robot having a maximum velocity of 0.3$\,$m/s.  As in the simulated version of this case, we check the goal distance at multiple temporal points in the future to reduce oscillatory behavior.
\item \emph{3-Robot Circle}:
Three robots simultaneously try to move to antipodal points on a circle.
\end{itemize}
In all of our physical robot experiments, we left the maximum linear velocity at 0.3$\,$ m/s, but reduced the maximum rotational velocity to be 0.5$\,$ rad/s in order to match the actual limits of the robots used in the experiments. Results from these experiments can be seen in the companion video. 

\begin{figure}[t]
\centering
\captionsetup[subfigure]{justification=centering}
\subfloat[3-Robot Circle]
{\includegraphics[width=0.7\columnwidth]{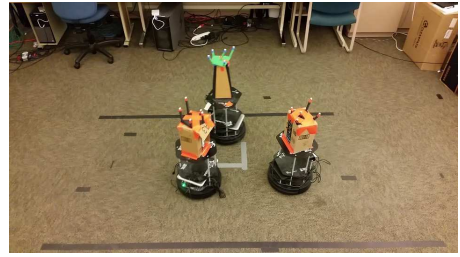}}
\quad
\vspace{0.2pt}

\subfloat[Robots' Controls]
{\includegraphics[width=0.8\columnwidth]{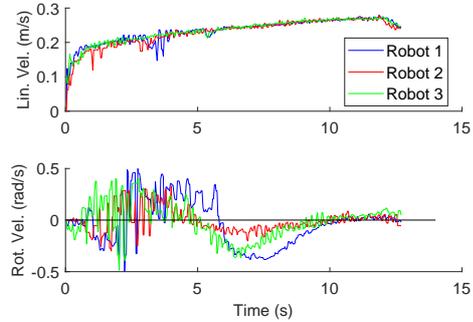}}
\caption{\textbf{Physical Robots - 3-Robot Circle}: Three robots avoid each other while moving to antipodal positions on a circle. 
The controls executed by each robot, as measured by wheel encoders and gyroscope, are shown in (b), illustrating the ability of the differential drives to smoothly adapt their linear and angular velocities to resolve impending collisions.
}
\label{fig:real_robot_circle}
\end{figure}

To further study the quality of the generated trajectories, we plotted the controls taken by the robot over time, as measured by its wheel encoders and gyroscope, in Figures~\ref{fig:real_robot_lane} and \ref{fig:real_robot_circle}. While there is some inherent noise in the measured controls, the robots were able to smoothly achieve their goals without colliding, taking admissible controls that stayed within the given limits in all scenarios tested.

In the car-following scenario, while in motion, the average linear acceleration of the controlled robot was 0.008$\,\mathrm{m/s^2}$ and its average rotational acceleration was 0.002$\,\mathrm{rad/s^2}$. In the first~8$\,$s, the robot gradually adapts its linear and angular speed 
to smoothly slip between the two obstacles.  Then, in the next~7$\,$s, it starts decelerating in order to align itself with the speed and orientation of the leader obstacle, after which it maintains an almost zero linear and angular acceleration.

In the 3-robot circle scenario, averaged across the times when the three differential drive robots were moving, the average linear acceleration was 0.008$\,\mathrm{m/s^2}$ and the average rotational acceleration was 0.011$\,\mathrm{rad/s^2}$.
Here, the robots are able to quickly resolve the collisions within the first 5$\,$s, and then smoothly adapt their orientations and accelerate to reach their goals. 
It is worth noting that the quick rotational velocity changes that the robots exhibit during the first 4$\,$s is due to the symmetric nature of the scenario; the robots are on track to arrive in the center of the environment nearly at the same time, and attempt to break the symmetry by trying to implicitly agree on whether to perform clockwise or counter clockwise avoidance maneuvers.

\subsection{Performance Analysis}
\begin{figure}
    \centering
    \includegraphics[width=0.9\columnwidth]{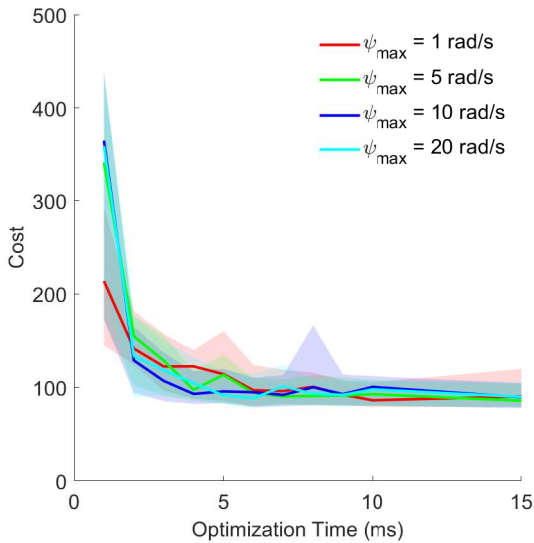}
    \caption{\textbf{Trajectory Cost Over Optimization Time}: 
    Performance analysis of NH-TTC for
    various control bounds in the car-following scenario shown in Figure~\ref{fig:overtake}.  The plot depicts the sum of the costs of the taken controls  as a function of the available planning time for a variety of maximum steering angle velocities, $\psi$. 
    Solid lines denote averages over 1,000 runs, with each run lasting 300 frames, and 
    shaded regions denote 90\% confidence intervals. 
    For a large variety of bounds, our approach is able to quickly find locally optimal solutions exhibiting low variance.}
    \label{fig:opt_cmp}
\end{figure}

We analyze the performance of our NH-TTC approach in the car-following scenario (see Section~\ref{sec:dyn_goals}) by varying the time that the robot has at its disposal to plan for a new control, as well as the maximum steering angle velocity that the car can attain. 
Figure~\ref{fig:opt_cmp} reports the total cost of the controlled-robot averaged over 1,000 runs for various planning times, ranging from 1$\,$ms to 15$\,$ms, and four distinct control bounds.
Overall, as can be seen in the figure, our subgradient descent implementation requires only a small optimization time to start finding low cost trajectories. 
Using, for example, a very tight constraint of 1$\,$rad/s on the steering angle, NH-TTC is able to find near-optimal solutions within 5$\,$ms of planning per time step. As we increase the control bounds from 1 $\,$rad/s to 20$\,$rad/s, the quality of the trajectories returned by NH-TTC remains nearly unchanged while the cost obtained still exhibits low variance across different runs. This highlights the ability of our approach to efficiently search the control space regardless of its size.  
In contrast, a sampling-based control approach would require more and more time to find good trajectories as the range of the robot's valid steering angle increases, making it impractical for real-time planning~settings.

\begin{figure*}
\captionsetup[subfigure]{justification=centering}
    \centering
    \null\hfill
    \subfloat[Scenario]
        {\fbox{\includegraphics[width=0.5\columnwidth,valign=c]{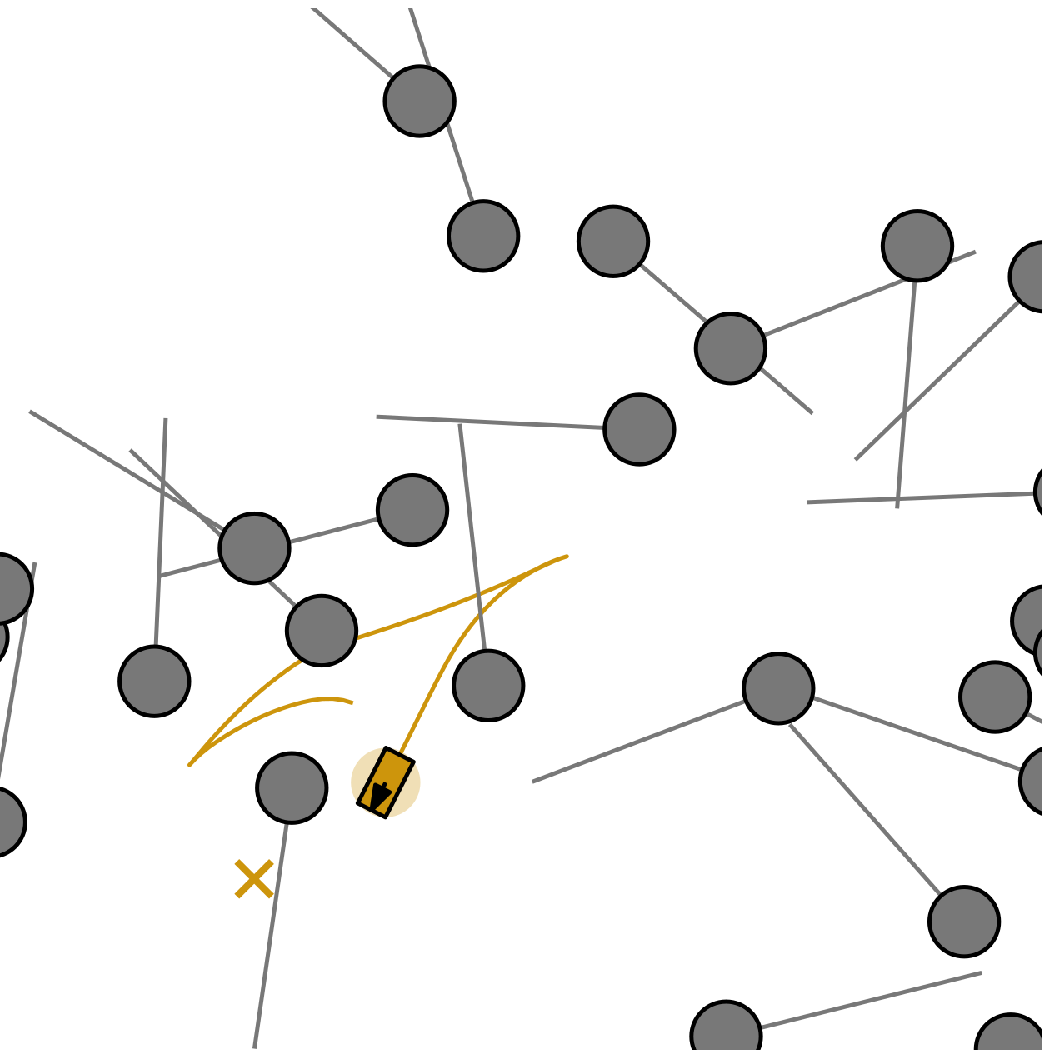}}}
    \hfill
    \subfloat[Control Cost Over Time]
        {\includegraphics[width=0.6\columnwidth,valign=c]{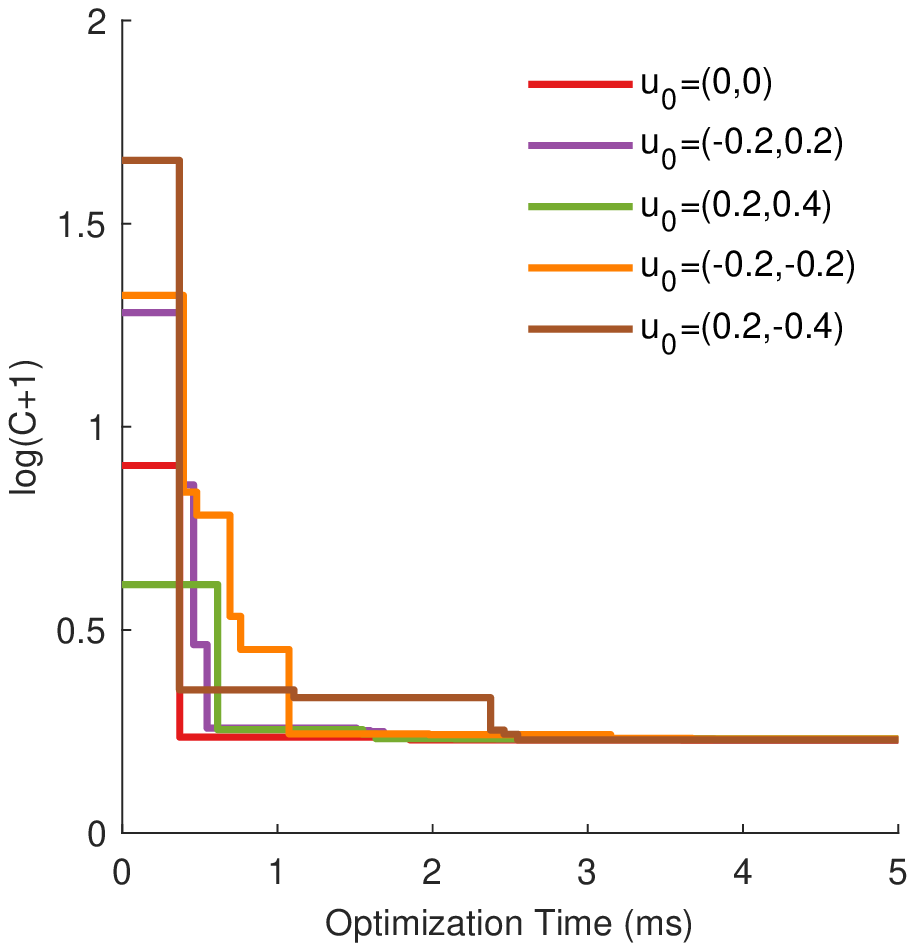}}
    \hfill
    \subfloat[Cost Field]
        {\includegraphics[width=0.6\columnwidth,valign=c]{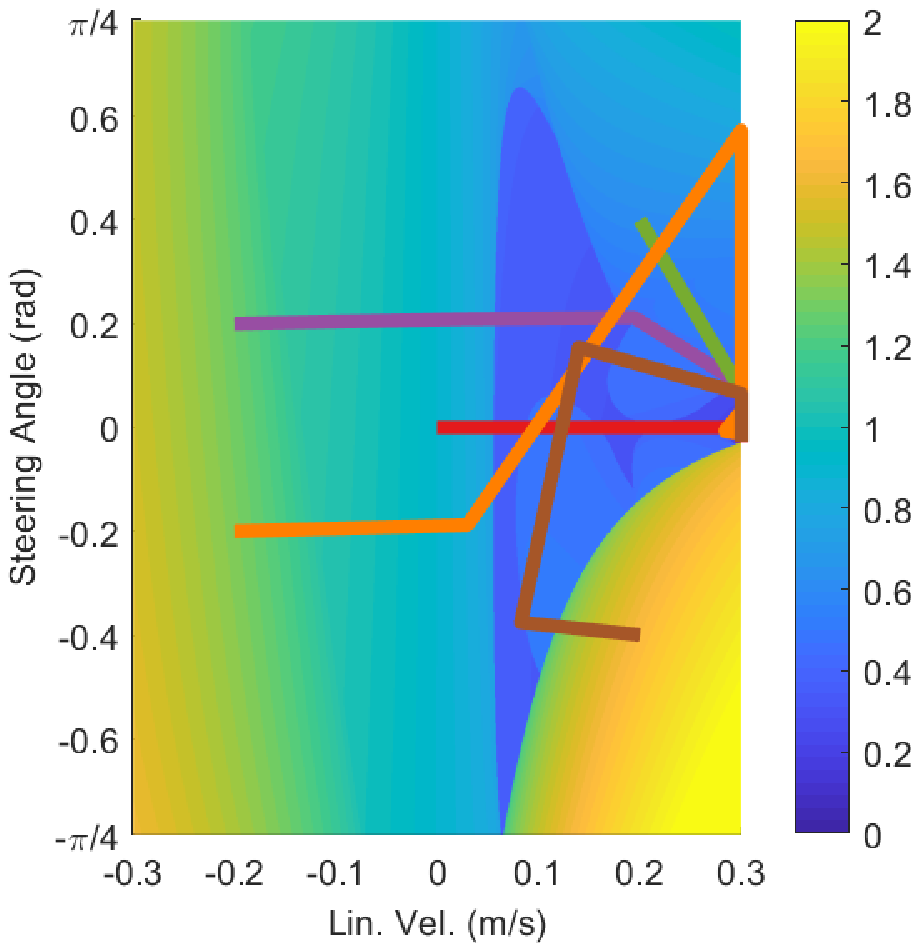}}
    \hfill\null
    \caption{\textbf{Effect of Initial Control Guess}: (a) A simple car navigating to the goal (the yellow X), avoiding the random velocity agents. (b) The log cost of the best control found so far for a variety of initializations over 5\,ms of planning. (c) The cost field corresponding to the simple car's controls.  To better show the gradient, values are plotted in log scale with the color corresponding to $\log(C(\mathbf{u}) + 1)$.  The evolution of each optimization run is shown in its corresponding color.
    }
    \label{fig:rand_car_cost2}
\end{figure*}

To further show the  robustness of our subgradient-based optimization, we test its 
sensitivity to the initial control,  $\mathbf{u}_0$, given as input to Algorithm~\ref{alg:sgd}. 
In particular, we chose a snapshot from the random scenario shown in 
Figure~\ref{fig:rand_car_cost2}a, where a simple car-like robot is interacting closely with a large number of dynamic obstacles. We ran NH-TTC using five different initial controls while allowing 5$\,$ms of planning time for the car. 
Across all five runs, NH-TTC completes between 164 and 185 subgradient descent iterations within the 5$\,$ms of the given planning time. 
Figures~\ref{fig:rand_car_cost2}b-c show the evolution over time of the best cost seen so far  and the corresponding control for each of the runs, where the 
first descent iteration is delayed by 0.3$\,$ms to pre-compute the obstacle trajectories.
As highlighted by the cost field in  Figure~\ref{fig:rand_car_cost2}c, the car has to solve a complex optimization problem, with many local minima. However, while the initial costs across all of the runs has a 
large spread, NH-TTC is able to quickly converge to similar near-optimal solutions in only~3\,ms.

\section{Conclusion} \label{sec:conclusion}

In this paper we have proposed NH-TTC, a new generalizable framework for anticipatory collision avoidance.  Our method is able to optimize directly in the control space of the robot in an anytime fashion, allowing collision-free trajectories to be computed over very short planning times for a wide variety of robot motion models.
To do so, we cast local navigation as a control optimization problem and employ an anticipatory cost function that focuses on the expected future values of robot controls. As such a function is non-convex and suffers from discontinuities, we minimize it using subgradient descent, and use implicit differentiation to capture the dynamics of future collisions for arbitrary motion models.

In an essence, our approach can be considered as a hybrid between full trajectory optimization methods and local reactive planning methods. Similar to trajectory optimization approaches, we decompose our cost function into two terms: a time-to-collision based trajectory cost, which in our case is discounted based on the risk of future collisions, and a cost-to-goal term that  accounts for the goal progress of the robot. However, analogous to local navigation methods, we define our trajectory by a single control propagated over time, allowing us to quickly resolve collisions with obstacles and further refine our solution in an anytime fashion. To the best of our knowledge, NH-TTC is the first ``local" approach that
demonstrates high-quality trajectories on a variety of kinematically constrained motion models in 
realtime and reactive settings.


\textit{Limitations:} While our approach performs well in many scenarios, there are certain types of motion models and scenarios we cannot properly address.  Since we are optimizing a single control, we are unable to operate on any unstable systems (such as a humanoid robot) where a single control cannot be taken over long horizons.  In addition, this single control optimization limits the type of maneuvers we can generate, as we greedily optimize goal distance in addition to finding collision-free controls. Finally, our Polyak-based step size of the search direction (Equation~\ref{eq_poly}) is domain-agnostic, as most of the times we do not know the optimal cost and we have to rely on the current best estimate.
This inefficiency, along with the optimizations discussed in Section~\ref{sec:implementation}, could be improved upon if the dynamics model of the robot is known a priori.


\textit{Future Work:}
We are excited to test the application of NH-TTC to other mobile robot types, especially those having 3D dynamics  such as quadrotors. In the future, we would also like to extend NH-TTC to account for motion and sensing uncertainty in the future trajectories of obstacles. Prior work on uncertainty-aware local navigation~\citep{calu,uttc} can provide some interesting ideas in this direction.
Furthermore, we would like to relax some of the assumptions that our framework makes, such as that the robot will maintain a constant control input over a finite time horizon.
Finally, we currently fit a single collision disk around each robot. Even though the center of the disk is not necessarily tied to the position of the robot's state, we may still underestimate the true time-to-collision value between a robot and a given obstacle, which can lead to conservative avoidance maneuvers (e.g., when simulating a simple car-like robot). To address this issue we would like to better approximate the geometry of the robot. A simple approach is to wrap a sequence of disks around the true shape of the robot and determine the minimal time to collision between each disk and a given obstacle. A better alternative may be to compute a tight fitting bounding shape for a robot using the medial axis transform as in~\cite{ctmat}. 

\section*{Funding}
This work was supported in part by the National Science Foundation under Grants IIS-1748541 and CNS-1544887.

\bibliographystyle{SageH}
\bibliography{references}


\end{document}